# Detection and Resolution of Rumours in Social Media: A Survey

ARKAITZ ZUBIAGA, University of Warwick, UK
AHMET AKER, University of Sheffield, UK and University of Duisburg-Essen, Germany
KALINA BONTCHEVA, University of Sheffield, UK
MARIA LIAKATA and ROB PROCTER, University of Warwick and Alan Turing Institute, UK

Despite the increasing use of social media platforms for information and news gathering, its unmoderated nature often leads to the emergence and spread of rumours, i.e., items of information that are unverified at the time of posting. At the same time, the openness of social media platforms provides opportunities to study how users share and discuss rumours, and to explore how to automatically assess their veracity, using natural language processing and data mining techniques. In this article, we introduce and discuss two types of rumours that circulate on social media: long-standing rumours that circulate for long periods of time, and newly emerging rumours spawned during fast-paced events such as breaking news, where reports are released piecemeal and often with an unverified status in their early stages. We provide an overview of research into social media rumours with the ultimate goal of developing a rumour classification system that consists of four components: rumour detection, rumour tracking, rumour stance classification, and rumour veracity classification. We delve into the approaches presented in the scientific literature for the development of each of these four components. We summarise the efforts and achievements so far toward the development of rumour classification systems and conclude with suggestions for avenues for future research in social media mining for the detection and resolution of rumours.

**ACM Reference format:**
Arkaitz Zubiaga, Ahmet Aker, Kalina Bontcheva, Maria Liakata, and Rob Procter. 2018. Detection and Resolution of Rumours in Social Media: A Survey. *ACM Comput. Surv.* 51, 2, Article 32 (February 2018), 36 pages.
https://doi.org/10.1145/3161603

This article has been supported by the PHEME FP7 Project (Grant No. 611233), the EPSRC Career Acceleration Fellowship No. EP/I004327/1, the COMRADES H2020 Project (Grant No. 687847), and the SoBigData Research Infrastructure (Grant No. 654024). We wish to thank the Alan Turing Institute for its support.
Authors' addresses: A. Zubiaga, M. Liakata, and R. Procter, University of Warwick, Department of Computer Science, Gibbet Hill Road, Coventry CV4 7AL, United Kingdom; emails: {a.zubiaga, m.liakata, rob.procter}@warwick.ac.uk; A. Aker, Universität Duisburg-Essen, Campus Duisburg, Arbeitsgruppe "Informationssysteme", Fakultät für Ingenieurwissenschaften, Abt. Informatik und Angew. Kognitionswissenschaft, 47048 Duisburg, Germany; email: a.aker@is.inf.uni-due.de; K. Bontcheva, University of Sheffield, Department of Computer Science, Regent Court, 211 Portobello, Sheffield S1 4DP, United Kingdom; email: k.bontcheva@sheffield.ac.uk.

# 1 INTRODUCTION

Social media platforms are increasingly being used as a tool for gathering information about, for example, societal issues (Lazer et al. 2009), and to find out about the latest developments during breaking news stories (Phuvipadawat and Murata 2010). This is possible because these platforms enable anyone with an internet-connected device to share in real-time their thoughts and/or to post an update about an unfolding event that they may be witnessing. Hence, social media has become a powerful tool for journalists (Diakopoulos et al. 2012; Tolmie et al. 2017) but also for ordinary citizens (Hermida 2010). However, while social media provides access to an unprecedented source of information, the absence of systematic efforts by platforms to moderate posts also leads to the spread of misinformation (Procter et al. 2013b; Webb et al. 2016), which then requires extra effort to establish their provenance and veracity. Updates associated with breaking news stories are often released piecemeal, which gives rise to a significant proportion of those updates being unverified at the time of posting, some of which may later be proven to be false (Silverman 2015a). In the absence of an authoritative statement corroborating or debunking an ongoing rumour, it is observed that social media users will often share their own thoughts on its veracity via a process of collective, inter-subjective sense-making (Tolmie et al. 2018) that may lead to the exposure of the truth behind the rumour (Procter et al. 2013a; Li and Sakamoto 2015).

Nevertheless, despite this apparent robustness of social media, its increasing tendency to give rise to rumours motivates the development of systems that, by gathering and analysing the collective judgements of users (Lukasik et al. 2016), are able to reduce the spread of rumours by accelerating the sense-making process (Derczynski and Bontcheva 2014). A rumour detection system that identifies, in its early stages, postings whose veracity status is uncertain, can be effectively used to warn users that the information in them may turn out to be false (Zhao et al. 2015). Likewise, a rumour classification system that aggregates the evolving, collective judgements posted by users can help track the veracity status of a rumour as it is exposed to this process of collective sense-making (Metaxas et al. 2015). In this article, we present an overview of the components needed to develop such a rumour classification system and discuss the success so far of the efforts toward building it.

## 1.1 Defining and Characterising Rumours

*Rumour Definition.* Recent publications in the research literature have used definitions of rumours that differ from one another. For example, some recent work has misdefined a rumour as an item of information that is deemed false (e.g., Cai et al. (2014) and Liang et al. (2015)), while the majority of the literature defines rumours instead as "unverified and instrumentally relevant information statements in circulation" (DiFonzo and Bordia 2007). In our article, we have adopted as the defining characteristic of rumours that they are unverified at the time of posting, which is consistent with the definition given by major dictionaries, such as the Oxford English Dictionary, which defines a rumour as "a currently circulating story or report of uncertain or doubtful truth"[1] or the Merriam Webster Dictionary, which defines it as "a statement or report current without known authority for its truth"[2]. This unverified information may turn out to be true, or partly or entirely false; alternatively, it may also remain unresolved. Hence, throughout this article, we adhere to this prevailing definition of rumour that classifies it as "*an item of circulating information whose veracity status is yet to be verified at the time of posting.*" The choice of this definition differs from some of the recent literature in social media research; however, it is consistent with major dictionaries and a longstanding research area in the social sciences (Allport and Postman 1946; Donovan 2007). A rumour can be understood as an item of information that has not yet been verified, and hence its truth value remains unresolved while it is circulating. A rumour is defined as unverified when there is no evidence supporting it or there is no official confirmation from authoritative sources (e.g., those with a reputation for being trustworthy) or sources that may have credibility in a particular context (e.g., eyewitnesses).

---

[1]https://en.oxforddictionaries.com/definition/rumour.
[2]http://www.merriam-webster.com/dictionary/rumor.

*Rumour Types.* Many different factors are available for classifying rumours by type, including its eventual veracity value (true, false, or unresolved) (Zubiaga et al. 2016c) or its degree of credibility (e.g., high or low) (Jaeger et al. 1980). Another attempt at classifying rumours by type is that by Knapp (1944), who introduced a taxonomy of three types of rumours: (1) "pipe-dream" rumours: i.e., rumours that lead to wishful thinking; (2) "bogy" rumours: i.e., those that increase anxiety or fear; and (3) "wedge-driving" rumours: i.e., those that generate hatred. When it comes to the development of a rumour classification system, the factor that largely determines approaches to be utilised is their temporal characteristics:

(1) *New rumours that emerge during breaking news.* Rumours that emerge in the context of breaking news are generally ones that have not been observed before. Therefore, rumours need to be automatically detected and a rumour classification system needs to be able to deal with new, unseen rumours, considering that the training data available to the system may differ from what will later be observed by it. In these cases, where early detection and resolution of rumours is crucial, a stream of posts needs to be processed in real-time. An example of a rumour that emerges during breaking news would be when the identity of a suspected terrorist is reported. A rumour classification system may have observed other similar cases of suspected terrorists, but the case and the names involved will most likely differ. Therefore, the design of a rumour classifier in these cases will need to consider the emergence of new cases, with the new vocabulary that they will likely bring.

(2) *Long-standing rumours that are discussed for long periods of time.* Some rumours may circulate for long periods of time without their veracity being established with certainty. These rumours provoke significant, ongoing interest, despite (or perhaps because of) the difficulty in establishing the actual truth. This is, for example, the case of the rumour stating that *Barack Obama is Muslim*. While this statement is unsubstantiated, it appears that there is no evidence that helps debunk it to the satisfaction of everyone.[3] For rumours like these, a rumour classification system may not need to detect the rumour, as it might be known *a priori*. Moreover, the system can make use of historical discussions about the rumour to classify ongoing discussions, where the vocabulary is much less likely to differ, and therefore the classifier built on old data can still be used for new data. In contrast to newly emerging rumours, for long-standing rumours, processing is usually retrospective, so posts do not necessarily need to be processed in real-time.

Throughout the article, we refer to these two types of rumours, describing how different approaches can deal with each of them.

### 1.2 Studying Rumours: From Early Studies to Social Media

*A Brief History.* Rumours and related phenomena have been studied from many different perspectives (Donovan 2007), ranging from psychological studies (Rosnow and Foster 2005) to computational analyses (Qazvinian et al. 2011). Traditionally, it has been very difficult to study people's reactions to rumours, given that this would involve real-time collection of reaction as rumours unfold, assuming that participants had already been recruited. To overcome this obstacle, Allport (Allport and Postman 1946, 1947) undertook early investigations in the context of wartime rumours. He posited the importance of studying rumours, emphasising that "newsworthy events are likely to breed rumors" and that "the amount of rumor in circulation will vary with the importance of the subject to the individuals involved times the ambiguity of the evidence pertaining to the topic at issue.' This led him to set forth a motivational question that is yet to be answered: "Can rumors be scientifically understood and controlled?" (Allport and Postman 1946). His 1947 experiment (Allport and Postman 1947) reveals an interesting fact about rumour circulation and belief. He looked at how U.S. President Franklin D. Roosevelt allayed rumours about losses sustained by the U.S. Navy at the Japanese attack on Pearl Harbor in 1941. The study showed that before the President made his address, 69% of a group of undergraduate students believed that losses were greater than officially stated; but five days later, the President having spoken in the meantime, only 46% of an equivalent group of students believed this statement to be true. This study revealed the importance of an official announcement by a reputable person in shaping society's perception of the accuracy of a rumour.

---

[3]Arguably, such rumours survive because they are a vehicle for those inclined to believe in conspiracy theories, where, by definition, nothing is as it seems.

Early research focused on different objectives. Some work has looked at the factors that determine the diffusion of a rumour, including, for instance, the influence of the believability of a rumour on its subsequent circulation, where believability refers to the extent to which a rumour is likely to be perceived as truthful. Early research by Prasad (1935) and Sinha (1952) posited that believability was not a factor affecting rumour mongering in the context of natural disasters. More recently, however, Jaeger et al. (1980) found that rumours were passed on more frequently when the believability level was high. Moreover, Jaeger et al. (1980) and Scanlon (1977) found the importance of a rumour as perceived by recipients to be a factor that determines whether or not it is spread, the least important rumours being spread more.

*Rumours on the Internet.* The widespread adoption of the internet gave rise to a new phase in the study of rumour in naturalistic settings (Bordia 1996) and has taken on particular importance with the advent of social media, which not only provides powerful new tools for sharing information but also facilitates data collection from large numbers of participants. For instance, Takayasu et al. (2015) used social media to study the diffusion of a rumour circulating during the 2011 Japan earthquake, which stated that rain in the aftermath might include harmful chemical substances and led to people being warned to carry an umbrella. The authors looked at retweets (RTs) of early tweets reporting the rumour, as well as later tweets reporting that it was false. While their study showed that the appearance of later correction tweets diminished the diffusion of tweets reporting the false rumour, the analysis was limited to a single rumour and does not provide sufficient insight into understanding the nature of rumours in social media. Their case study, however, does show an example of a rumour with important consequences for society, as citizens were following the latest updates with respect to the earthquake to stay safe.

*Rumours in Social Media.* Social media as a source for researching rumours has gained ground in recent years, both because it is an interesting source for gathering large datasets associated with rumours and also because, among other factors, its huge user base and ease of sharing makes it a fertile breeding ground for rumours. Research has generally found that Twitter does well in debunking inaccurate information thanks to self-correcting properties of crowdsourcing as users share opinions, conjecture,s and evidence. For example, Castillo et al. (2013) found that the ratio between tweets supporting and debunking false rumours was 1:1 (one supporting tweet per debunking tweet) in the case of a 2010 earthquake in Chile. Procter et al. (2013b) came to similar conclusions in their analysis of false rumours during the 2011 riots in England, but they noted that any self-correction can be slow to take effect. In contrast, in their study of the 2013 Boston Marathon bombings, Starbird et al. (2014) found that Twitter users did not do so well in distinguishing between the truth and hoaxes. Examining three different rumours, they found the equivalent ratio to be 44:1, 18:1, and 5:1 in favour of tweets supporting false rumours. Delving further into temporal aspects of rumour diffusion and support, Zubiaga et al. (2016c) describe the analysis of rumours circulating during nine breaking news events. This study concludes that, while the overall tendency is for users to support unverified rumours in the early stages, there is a shift toward supporting true rumours and debunking false rumours as time goes on. The ability of social media to aggregate the judgements of a large community of users (Li and Sakamoto 2015) thus motivates further study of machine learning approaches to improve rumour classification systems. Despite the challenges that the spread of rumours and misinformation pose for the development of such systems, breaking down the development process into smaller components and making use of suitable techniques is showing encouraging progress toward developing effective systems that can assist people in making decisions towards assessing the veracity of information gathered from social media.

### 1.3 Scope and Organisation

This survey article is motivated by the increasing use of social media platforms such as Facebook or Twitter to post and discover information. While we acknowledge their unquestionable usefulness for gathering often exclusive information, their openness, lack of moderation, and the ease with which information can be posted from anywhere and at any time undoubtedly leads to major problems for information quality assurance. Given the unease that the spread of rumours can produce and the potential for harm, the incentive for the development of data mining tools for dealing with rumours has increased in recent years. This survey article aims to delve into these challenges posed by rumours to the development of data mining applications for gathering information from social media, as well as to summarise the efforts so far in this direction.

We continue this survey in Section 2 by examining the opportunities social media brings to numerous domains, while also introducing the new challenge of having to deal with rumours. Moving on to the analysis of rumour classification systems, we first describe different approaches to putting together a dataset of rumours that enables further experimentation; the generation of datasets is described in Section 3, beginning with ways for accessing social media APIs, to outlining approaches for collecting and annotating data collected from social media. We summarise findings from studies looking at the characterisation and understanding of diffusion and dynamics of rumours in social media in Section 4. After that, we describe the components that form a rumour classification system in Section 5. These components are then further described and existing approaches discussed in subsequent sections; rumour detection systems in Section 6, rumour tracking systems in Section 7, rumour stance classification in Section 8, and veracity classification in Section 9. We continue in Section 10 listing and describing existing applications that deal with the classification of rumours and related applications. To conclude, we summarise the achievements to date and outline future research directions in Section 11.

## 2 SOCIAL MEDIA AS AN INFORMATION SOURCE: CHALLENGES POSED BY RUMOURS

Social media is being increasingly leveraged by both a range of professionals as well as members of the public as an information source to learn about the latest developments and current affairs (Van Dijck 2013; Fuchs 2013). The use of social media has been found useful in numerous different domains; we describe some of the most notable uses below:

*News Gathering*. Social media platforms have shown great potential for news diffusion, occasionally even outpacing professional news outlets in breaking news reporting (Kwak et al. 2010). This enables, among others, access to updates from eyewitnesses and a broad range of users who have access to potentially exclusive information (Diakopoulos et al. 2012; Starbird et al. 2012). Aiming to exploit this feature of social media platforms, researchers have looked into the development of tools for news gathering (Zubiaga et al. 2013; Diakopoulos et al. 2012; Marcus et al. 2011), analysed the use of user-generated content (UGC) for news reporting (Hermida and Thurman 2008; Tolmie et al. 2017), and explored the potential of social media to give rise to collaborative and citizen journalism, including collaborative verification of reports posted in social media (Hermida 2012; Spangenberg and Heise 2014).

*Emergencies and Crises.* The use of social media during emergencies and crises has also increased substantially in recent years (Imran et al. 2015; Castillo 2016; Procter et al. 2013a), with applications such as getting reports from eyewitnesses or finding those seeking help. Social media has been found useful for information gathering and coordination in different situations, including emergencies (Yates and Paquette 2011; Yin et al. 2012; Procter et al. 2013a), protests (Trottier and Fuchs 2014; Agarwal et al. 2014), and natural hazards (Vieweg et al. 2010; Middleton et al. 2014).

*Public Opinion.* Social media is also being used by researchers to collect perceptions of users on a range of social issues, which can then be aggregated to measure public opinion (Murphy et al. 2014). Researchers attempt to clean social media data (Gao et al. 2014) and try to get rid of population biases (Olteanu et al. 2016) to understand how social media shapes society's perceptions on issues, products, people, and the like. Goodman et al. (2011). Social media has been found useful for measuring public opinion during elections (Anstead and O'Loughlin 2015), and the effect of online opinions on, for instance, the reputation of organisations (Sung and Lee 2015) or attitudes toward health programmes (Shi et al. 2014).

*Financial/Stock Markets.* Social media has also become an important information source for staying abreast of the latest developments in the financial world and in stock markets. For instance, sentiment expressed in tweets has been used to predict stock market reactions (Azar and Lo 2016), to collect opinions that investors post in social media (Chen et al. 2014) or to analyse the effect that social media posts can have on brands and products (Lee et al. 2015).

Because of the increasing potential of social media as an information source, its propensity for the spreading of misinformation and unsubstantiated claims has given rise to numerous studies. Studies have looked at credibility perceptions of users (Westerman et al. 2014) and have also assessed the degree to which users rely on social media to gather information such as news (Gottfried and Shearer 2016). The difficulties arising from the presence of rumours and questionable claims in social media has hence led to interest in techniques for building rumour classification systems and to alleviate the problem by facilitating the gathering of accurate information for users. When it comes to the development of rumour classification systems, there are two main use cases to be considered:

— *Dealing with Long-Standing Rumours.* Where the rumours being tracked are known *a priori* and social media is being mined as a source for collecting opinions. This use case may be applicable, for instance, when wanting to track public opinion, or when rumours such as potential buyouts are being discussed for long periods in the financial domain.
— *Dealing with Emerging Rumours.* Where new rumours emerge suddenly while certain events or topics are being tracked. This use case may apply in the case of news gathering and emergencies, where information is released piecemeal and needs to be verified, or other suddenly emerging rumours, such as those anticipating political decisions that are expected to have an impact on stock markets.

## 3 DATA COLLECTION AND ANNOTATION

This section describes different strategies used to collect social media data that enables researching rumours, as well as approaches for collecting annotations for the data.

## 3.1 Access to Social Media APIs

The best way to access, collect, and store data from social media platforms is generally through application programming interfaces (APIs) (Lomborg and Bechmann 2014). APIs are easy-to-use interfaces that are usually accompanied by documentation that describes how to request the data of interest. They are designed to be accessed by other applications as opposed to web interfaces, which are designed for people; APIs provide a set of well-defined methods that an application can invoke to request data. For instance, in a social media platform, it may be desirable to retrieve all data posted by a specific user or all the posts containing a certain keyword.

Before using an API, a crucial first step is to read its documentation and to understand its methods and limitations. Indeed, every social media platform has its own limitations and this is key when wanting to develop a rumour classification system that utilises social media data. Three of the key platforms used for the study of rumours are Twitter, Sina Weibo, and Facebook; here we briefly discuss the features and limitations of these three platforms:

—Twitter provides detailed documentation[4] of ways to use its API, which gives access to a REST API to harvest data from its database as well as a streaming API to harvest data in real-time. After registering a Twitter application[5] that will generate a set of keys for accessing the API through OAuth authentication, the developer will then have access to a range of methods ("endpoints") to collect Twitter data. The most generous of these endpoints gives access to a randomly sampled 1% of the whole tweet stream; getting access to a larger percentage usually requires payment of a fee. To make sure that a comprehensive collection of tweets has been gathered, it is advisable to collect tweets in real-time through the streaming API; again, there is a limit of 1% on the number of tweets that can be collected for free from this API. The main advantage of using Twitter's API is that it is the most open and this may partly explain why it is the most widely used for research; the main caveat is that it is mainly designed to collect either real-time or recent data, and so it is more challenging to collect data that is older than the last few weeks. Twitter provides a range of metadata with each tweet collected, including tweet language, location (where available), and so on, as well as details of the user posting the tweet.
—Sina Weibo, the most popular microblogging platform in China, provides an API[6] that has many similarities to that of Twitter. However, access to some of its methods is not openly available. For example the search API requires contacting the administrator to get approval first. Moreover, the range of methods provided by Sina Weibo are only accessible through its REST API and it lacks an official streaming API to retrieve real-time data. To retrieve real-time data from Sina Weibo through its streaming API, it is necessary to make use of third party providers such as Socialgist.[7,8] As with Twitter, Sina Weibo provides a set of metadata with each post, including information about the post and details of the user.
—Facebook provides a documented API[9] with a set of software development kits for multiple programming languages and platforms that make it easy to develop applications with its data. Similar to the Twitter API, Facebook also requires registering an application[10] to generate the keys needed to access the API. In contrast to Twitter, most of the content posted by Facebook users is private and therefore there is no access to specific content posted, unless the users are "friends" with the authenticated account. The workaround to get access to posts on Facebook is usually to collect data from so-called Facebook Pages, which are open pages created by organisations, governments, groups, or associations. Unlike Twitter, it is possible to then get access to historical data from those Facebook Pages; however, access is limited to content that has been posted in those pages. Metadata provided with each post is more limited with Facebook and requires additional requests to the API to get them.

---

[4]https://dev.twitter.com/docs.
[5]https://apps.twitter.com.
[6]http://open.weibo.com/wiki/API%E6%96%87%E6%A1%A3/en.
[7]http://www.socialgist.com/.
[8]http://www.socialgist.com/press/socialgist-emerges-as-the-first-official-provider-of-social-data-from-chinese-microblogging-platform-sina-weibo/.
[9]https://developers.facebook.com/docs/.
[10]https://developers.facebook.com/docs/apps/register.

When using these APIs, it is important to take notice of the potential impact of restrictions imposed by the Terms of Service of the platform in question, especially when it is intended to release a dataset publicly. These tend to change frequently and hence it is recommended to check them before releasing a dataset. Often, releasing raw data is not permitted and is limited to the unique identifiers of the content, e.g., tweet IDs; users must then use the IDs to collect raw the data themselves (Weller and Kinder-Kurlanda 2016). There are exceptions, however; Twitter, for instance, allows researchers to release small datasets of up to 50,000 tweets, including raw data.

In recent years, Twitter has become the data source *par excellence* for collection and analysis of rumours, thanks to the openness of its API, as well as its prominence as a source of early reports during breaking news. Most of the research surveyed in this study, as well as the applications described in Section 10, make use of Twitter for this reason.

### 3.2 Rumour Data Collection Strategies

Collection of social media data that is relevant for the development of rumour classifiers is not straightforward *a priori* and it is necessary to define a careful data collection strategy to come up with good datasets. For other applications in social media mining, it might just suffice to define filters that are already implemented in the APIs of social media platforms, such as: (1) filtering by keyword to collect data related to an event (Driscoll and Walker 2014); (2) defining a bounding box to collect data posted from predefined geographical locations (Frias-Martinez et al. 2012); or (3) listing a set of users of interest to track their posts (Li and Cardie 2014). Collection of rumours requires combining one of those implemented approaches with expertise to retrieve data that is applicable to the rumour classification scenario.

We classify the different data collection strategies employed in the literature on two different levels. On one hand, researchers have used different strategies to collect long-standing rumours or newly emerging rumours and, on the other, researchers have relied on different top-down and bottom-up strategies for sampling rumour-related data from their collections.

*Collection of long-standing rumours vs collection of emerging rumours.* The methodology for collecting rumour data from social media can have important differences, depending on whether the aim is to collect long-standing or newly emerging rumours.

— Collection of long-standing rumours is performed for a rumour or rumours that are known in advance. For instance, posts can be collected for the rumour discussing whether *Obama is Muslim or not* by using keywords like *Obama* and *muslim* to filter the posts (Qazvinian et al. 2011). Since such rumours have, by definition, been running for a while, there is no need to have a system that detects those rumours and the list of rumours is manually input. This type of collection is useful when wanting to track opinion shifts over a long period of time, and the ease with which keywords can be defined to collect posts enables collection of large-scale datasets. It is important to be careful when defining the keywords, so that as many relevant posts as possible are collected.
—Collection of emerging rumours tends to be more challenging. Given that data collection is usually done from a stream of posts in real-time, it is necessary to make sure that tweets associated with a rumour will be collected before it occurs. Since the keywords are not known beforehand, alternative solutions are generally used for performing a broader data collection to then sample the subset of interest. In closed scenarios where there is a need to make sure that rumours that emerge during an event or news story are collected, the simple approach is to collect as many posts as possible for those events. Once the posts for an event are collected, the tweets that are associated with rumours can then filtered (Procter et al. 2013b; Zubiaga et al. 2015). This can be done in two different ways by following top-down or bottom-up strategies, as we explain below. Alternatively, it may be desirable to collect emerging rumours in an open scenario that is not restricted to events or news stories, but in a broader context. A solution for this is to use alternative API endpoints to collect posts through a less restrictive stream of data, such as Twitter's streaming API sampling a random 1% of the whole, or a filter of posts by geolocation, where available, to collect posts coming from a country or region (Han et al. 2014). A caveat to be taken into account is that, since the data collection has not been specifically set up for a rumour but for a wider collection, the sampled subset associated with rumours may not lead to comprehensive representations of the rumours as keywords different to those initially predefined can be used. The identification of changes in vocabulary during an event for improved data collection is, however, an open research issue (Earle et al. 2012; Wang et al. 2015).

*Top-down vs. bottom-up data sampling strategies.* When performing a broad collection, for instance, as may be the case when attempting to discover newly emerging rumours, collecting posts related to an event or when following an unfiltered stream of posts, it may then be necessary to sample the data to extract the posts associated with rumours. This sampling can be performed by using either a top-down or a bottom-up strategy:

—Top-down sampling strategies prevailed in early work on social media rumours, i.e., sampling posts related to rumours identified in advance. This can apply to long-standing rumours, where keywords can be defined for sampling posts related to a rumour known to have been circulating for a long time (Qazvinian et al. 2011), for retrospective sampling of rumours known to have emerged during an event (Procter et al. 2013b), or using rumour repositories like Snopes.com to collect posts associated with those rumours (Hannak et al. 2014). The main caveat of this approach is that sampling of data is limited to the rumours listed and other rumours may be missed.
—Bottom-up sampling strategies have emerged more recently in studies that aimed at collecting a wider range of rumours, i.e., sifting through data to identify rumours, rather than rumours that are already known. Instead of listing a set of known rumours and filtering tweets related to those, the bottom-up collection consists of displaying a timeline of tweets so that an annotator can go through those tweets, identifying the ones that are deemed rumourous. This is an approach that was used first by Zubiaga et al. (2016c) and subsequently by Giasemidis et al. (2016). The benefit of this approach is that it leads to a wider range of rumours than the top-down strategy, as it is more likely to find new rumours that would not have been found otherwise. The main caveat of this approach is that generally leads to a few tweets associated with each rumour, rather than a comprehensive collection of tweets linked to each rumour as with the top-down strategy.

### 3.3 Annotation of Rumour Data

The annotation of rumour data can be carried out at different levels, depending on the task and the purpose. Here, we present previous efforts on rumour annotation for different purposes. The first step is to identify the rumourous subset within the collected data. This is sometimes straightforward as only rumourous data is collected using top-down sampling strategies, and hence no further annotation is needed to identify what is a rumour and what is not. However, when using a bottom-up sampling strategy, manual annotation work is needed to identify what constitutes a rumour and what a non-rumour (Zubiaga et al. 2015). Manual distinction of what is a rumour may not always be straightforward, as it is largely dependent on the context and on human judgement as to whether the underlying information was verified or not at the moment of posting. However, well-established definitions of rumours exist to help in this regard (DiFonzo and Bordia 2007) and people with a professional interest in veracity, such as journalists, have put together detailed guides to help determine what is a rumour (Silverman 2015a) and when further verification is needed (Silverman 2013). Annotation work distinguishing rumours and non-rumours is described in Zubiaga et al. (2016b) as a task to determine when a piece of information doesn't have sufficient evidence to be verified or lacks confirmation from an authoritative source.

Once rumours and non-rumours have been manually classified, further annotation is usually useful to do additional classification and resolution work on the rumours. It is usually the case that no additional annotations are collected for non-rumours, as the rumours are the ones that need to be further dealt with; different annotations that have been done on rumours include the following:

— *Rumour Veracity*: Manually determining the veracity of a rumour is challenging, usually requiring an annotator with expertise who performs careful analysis of claims and additional evidence, context and reports from authoritative sources before making a decision. This annotation process has been sometimes operationalised by enlisting the help of journalists with expertise in verification (Zubiaga et al. 2016c). In this example, journalists analysed rumourous claims spreading on social media during breaking news to determine, where possible, if a rumour had later been confirmed as true or debunked and proven false; this is, however, not always possible and some rumours were marked as *unverified* as no reliable resolution could be found. While this approach requires expertise that can be hard to resource, others have used online sources to determine the veracity of rumours. For instance, Hannak et al. (2014) used Snopes.com as a database that provides ground truth annotations of veracity for rumours put together by experts. While some online sources like Truth-O-Meter and PolitiFact provide finer-grained labels for veracity, such as *mostly true*, *half true* and *mostly false*, these are usually reduced to three labels (Popat et al. 2016), namely *true*, *false,* and optionally *unverified*. While annotation is increasingly being performed through crowdsourcing platforms for many natural language processing and data mining tasks (Doan et al. 2011; Wang et al. 2013), it is not as suitable for more challenging annotation tasks such as rumour veracity. Crowdsourcing annotations for veracity will lead to collection of credibility perceptions rather than ground truth veracity values, given that verification will often require an exhaustive work of checking additional sources for validating the accuracy of information, which may be beyond the expertise of average crowd workers. This is what Zubiaga and Ji (2014) found in their study, suggesting that verification work performed by crowd workers tends to favour selection of true labels for inaccurate information.
— *Stance Toward Rumours*: Typically, a rumour will provoke an exchange of views between social media users, with each post reflecting a particular *stance* on its likely veracity. These stances can be aggregated to help determine the veracity of the target rumour. Annotation of stance has been operationalised by Qazvinian et al. (2011) annotating tweets as *supporting*, *denying*, or *querying* a rumour, while later work by Procter et al. (2013b) suggested the inclusion of an additional label, *commenting*, to expand the annotation scheme to four labels. Posts are labelled as *supporting* or *denying* when they express a clear supporting or opposing stance toward the rumour; they are labelled as *querying* when a post questions the veracity of a rumour or appeals for more information; and as *commenting* when a post is either unrelated to the rumour or does not contribute in any way to the veracity of the rumour.

— *Rumour Relevance*: Annotation of relevance involves determining if a social media post is related to a rumour or not. This is operationalised as a binary annotation scheme, marking a post as either relevant or not. Qazvinian et al. (2011) annotated rumours by relevance where, for instance, for a rumour saying that *Obama is Muslim*, a post that says *Obama does seem to be Muslim* would be marked as relevant, while a post saying that *Obama had a meeting with Muslims* would be marked as not relevant.

— *Other Factors*: Some work has performed annotation of additional factors that can be of help in the assessment of the veracity of rumours. For example, Zubiaga et al. (2016c) annotated rumourous tweets for certainty (certain, somewhat certain, uncertain) and evidentiality (first-hand experience, inclusion of URL, quotation of person/organisation, link to an image, quotation of unverifiable source, employment of reasoning, no evidence), along with support. Lendvai et al. (2016a) annotated relations between claims associated with rumours, intended for automated identification of entailment and contradiction between claims. Annotation of credibility perceptions has also been done in previous work, determining how credible claims appear to people (Mitra and Gilbert 2015; Zubiaga and Ji 2014); this could be useful in the context of rumours to identify those that are likely to be misleading for people; however, it has not yet been applied in the context of rumours.

## 4 CHARACTERISING RUMOURS: UNDERSTANDING RUMOUR DIFFUSION AND FEATURES

Numerous recent studies have looked at characterising the emergence and spread of rumours in social media. Insights from these studies can, in turn, be useful to inform the development of rumour classification systems. Some of this research has focused on extensive analyses of a specific rumour, whereas others have looked into larger sets of rumours to perform broader analyses.

Studies of discourse surrounding rumours have been conducted to examine discussions around— and the evolution of—rumours over time. Some studies have looked at defining a scheme to categorise types of reactions expressed toward rumours. Maddock et al. (2015) looked at the origins and changes of rumours over time, which led to the identification of seven behavioural reactions to rumours: misinformation, speculation, correction, question, hedge, unrelated, or neutral/other. Similarly, Procter et al. (2013b) suggested that reactions to rumours could be categorised into four types, namely support, denial, appeal for more information, and comment. Others have looked into rumours to understand how people react to them. By looking at rumours spreading in the Chinese microblogging platform Sina Weibo, Liao and Shi (2013) identified interventions of seven types of users (celebrity, certified, mass media, organisation, website, internet star, and ordinary), who contributed in seven different ways (providing information, giving opinions, emotional statements, sense-making statements, interrogatory statements, directive statements, and digressive statements). In another study looking at "conversations" (i.e., a series of tweets linked by reply relationships) provoked by rumourous reports on Twitter, Zubiaga et al. (2016c) found that the prevalent tendency of social media users is to support and spread rumours, irrespective of their veracity value. This includes users of high reputation, such as news organisations, who tend to favour rumour support in the early stages of rumours, issuing a correction statement later where needed. In an earlier study, Mendoza et al. (2010) had found strong correlations between rumour support and veracity, showing that a majority of users support true rumours, while a higher number of users denies false rumours. Despite the apparent contradiction between these studies, it is worth noting that Mendoza et al. (2010) looked at the entire life cycle of a rumour and hence the aggregation leads to good correlations; in contrast, Zubiaga et al. (2016c) focused on the early reactions to rumours, showing that users have problems in determining veracity in the early stages of a rumour. Using rumour data from Reddit, differences across users have also been identified, suggesting three different user groups: those who generally support a false rumour, those who generally refute a false rumour, and those who generally joke about false rumours (Dang et al. 2016a). It has also been suggested that corrections are usually issued by news organisations and they can be sometimes widely spread (Takayasu et al. 2015; Arif et al. 2016; Andrews et al. 2016), especially if those corrections come from like-minded accounts (Hannak et al. 2014) and occasionally even leading to deletion or unsharing of the original post (Frias-Martinez et al. 2012). However, corrections do not always have the same effect as the original rumours (Lewandowsky et al. 2012; Shin et al. 2016; Starbird et al. 2014), which reinforces the need to develop rumour classification systems that deal with newly emerging rumours.

Other studies have looked at factors motivating the diffusion of rumours. Rumour diffusion is often dependent on the strength of ties between users, where rumours are more likely to be spread across strong ties in a network (Cheng et al. 2013). Other studies looking at temporal patterns of rumours have suggested that their popularity tends to fluctuate over time in social media (Kwon et al. 2013; Kwon and Cha 2014; Lukasik et al. 2015b) and other platforms on the internet (Jo 2002), but with a possibility of being discussed again later in time after rumour popularity fades.

Studies have also looked at the emergence of rumours. By using rumour theoretic approaches to examine factors that lead to expression of interest in tracking a rumour, Oh et al. (2013) identified the lack of an official source and personal involvement as the most important factors, whereas other factors, such as anxiety, were not as important. The poster's credibility and attractiveness of the rumour are also believed to be factors contributing to the propagation of rumours (Petty and Cacioppo 2012). Liu et al. (2014) reinforced these findings suggesting that personal involvement was the most important factor. Analysing specific rumour messages on Twitter, Chua et al. (2016) identified that tweets from established users with a larger follower network were spread the most.

While many studies have explored the diffusion of rumours, an exhaustive analysis of these studies is not within the scope of this survey article, which focuses instead on research concerning development of approaches to detect and resolve rumours. To read more about studies looking at the diffusion of rumours, we recommend the surveys by Serrano et al. (2015) and Walia and Bhatia (2016).

## 5 RUMOUR CLASSIFICATION: SYSTEM ARCHITECTURE

The architecture of a rumour classification system can have slight variations, depending on the specific use case. Here we define a typical architecture for a rumour classification system, which includes all the components needed for a complete system; however, as we point out in the descriptions below, depending on requirements, some of these components can be omitted. A rumour classification system usually begins with identifying that a piece of information is not confirmed (i.e., rumour detection) and ends by determining the estimated veracity value of that piece of information (i.e., veracity classification). The entire process from rumour detection to veracity classification is performed through the following four components (see Figure 1):

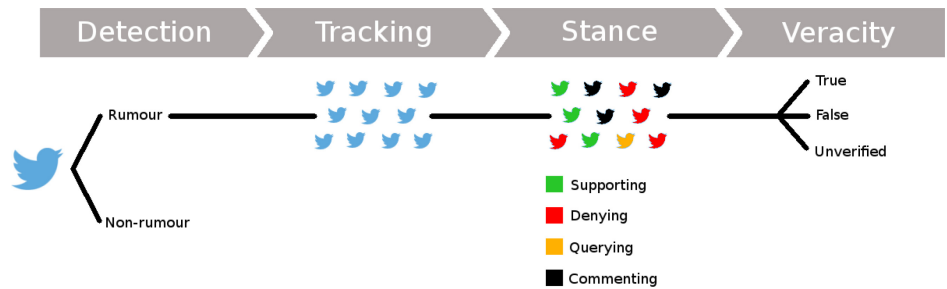

Fig. 1. Architecture of a rumour classification system.

(1) *Rumour detection*: In the first instance, a rumour classification system has to identify whether a piece of information constitutes a rumour. A typical input to a rumour detection component can be a stream of social media posts, whereupon a binary classifier has to determine if each post is deemed a rumour or a non-rumour. The output of this component is the stream of posts, where each post is labelled as rumour or non-rumour. This component is useful for identifying emerging rumours; however, it is not necessary when dealing with rumours that are known *a priori*.
(2) *Rumour tracking*: Once a rumour is identified, either because it is known *a priori* or because it is identified by the rumour detection component, the rumour tracking component collects and filters posts discussing the rumour. Having a rumour as input, which can be a post or a sentence describing it, or a set of keywords, this component monitors social media to find posts discussing the rumour, while eliminating irrelevant posts. The output of this component is a collection of posts discussing the rumour.
(3) *Stance classification*: While the rumour tracking component retrieves posts related to a rumour, the stance classification component determines how each post is orienting to the rumour's veracity. Having a set of posts associated with the same rumour as input, it outputs a label for each of those posts, where the labels are chosen from a generally predefined set of types of stances. This component can be useful to facilitate the task of the subsequent component dealing with veracity classification. However, it can be omitted where the stance of the public is not considered useful, e.g., cases solely relying on input from experts or validation from authoritative sources.
(4) *Veracity classification*: The final veracity classification component attempts to determine the actual truth value of the rumour. It can use as input the set of posts collected in the rumour tracking component, as well as the stance labels produced in the stance classification component. It can optionally try to collect additional data from other sources such as news media, or other websites and databases. The output of the component can be just the predicted truth value, but it can also include context such as URLs or other data sources that help the end user assess the reliability of the classifier by double checking with relevant sources.

In the following sections, we explore these four components in more detail, the approaches that have been used so far to implement them and the achievements to date.

## 6 RUMOUR DETECTION
### 6.1 Definition of the Task and Evaluation
The rumour detection task is that in which a system has to determine, from a set of social media posts, which ones are reporting rumours, and hence are spreading information that is yet to be verified. Note that the fact that a tweet constitutes a rumour does not imply that it will later be deemed true or false, but instead that it is unverified at the time of posting. Formally, the task takes a timeline of social media posts $TL = \{t_1, \ldots, t_{|TL|}\}$ as input, and the classifier has to determine whether each of these posts, $t_i$, is a rumour or a non-rumour by assigning a label from $Y = \{R, NR\}$. Hence, the task is usually formulated as a binary classification problem, whose performance is evaluated by computing the precision, recall, and F1 scores for the target category, i.e., rumours.

### 6.2 Datasets

The only publicly available dataset is the PHEME dataset of rumours and non-rumours,[11] which includes a collection of 1,972 rumours and 3,830 non-rumours associated with 5 breaking news stories (Zubiaga et al. 2016b).

### 6.3 Approaches to Rumour Detection

Despite the increasing interest in analysing rumours in social media and building tools to deal with rumours that had been previously identified (Seo et al. 2012; Takahashi and Igata 2012), there has been very little work in automatic rumour detection. Some of the work in rumour detection (Qazvinian et al. 2011; Hamidian and Diab 2015, 2016) has been limited to finding rumours known *a priori*. They feed a classifier with a set of predefined rumours (e.g., *Obama is Muslim*), which classifies new tweets as being related to one of the known rumours or not (e.g., *I think Obama is not Muslim* would be about the rumour, while *Obama was talking to a group of Muslims* would not). An approach like this can be useful for long-standing rumours, where what is required is to identify tweets relevant for tracking the rumours that have already been identified; in this survey article, we refer to this task as *rumour tracking*, as the rumours being monitored are known, but the stream of posts needs to be filtered. Relying solely on *rumour tracking* would not suffice for fast-paced contexts such as breaking news, where new, unseen rumours emerge and the specific keywords linked to a rumour, which is yet to be detected, are not known *a priori*. To deal with this, a classifier will need to learn generalisable patterns that will help identify rumours during emerging events.

The first work that tackled the detection of new rumours is that by Zhao et al. (2015). Their approach builds on the assumption that rumours will provoke tweets from skeptic users who question or enquire about their veracity; the fact that a piece of information has a number of enquiring tweets associated would then imply that the information is rumourous. The authors created a manually curated list of five regular expressions (e.g., "is (that | this | it) true") that are used to identify enquiring tweets. These enquiring tweets are then clustered by similarity, each cluster being ultimately deemed a candidate rumour. It was not viable for them to evaluate by recall, and instead only evaluated by precision.

In contrast, Zubiaga et al. (2016b, 2017) suggested an alternative approach that learns context throughout a breaking news story to determine if a tweet constitutes a rumour. They build on the hypothesis that a tweet alone may not suffice to know if its underlying story is a rumour, due to the lack of context. Moreover, they avoid the reliance on enquiring tweets, which they argue that not all rumours may trigger and hence may lead to low recall, as rumours not provoking enquiring tweets would be missed. Their context-learning approach relied on conditional random fields (CRF) as a sequential classifier that learns the reporting dynamics during an event, so that the classifier can determine, for each new tweet, whether it is or not a rumour based on what has been seen so far during the event. Their approach led to improved performance over the baseline classifier by Zhao et al. (2015), improving also a number of non-sequential classifiers compared as baselines. The classifier was also evaluated in terms of recall in this case, achieving state-of-the-art results.

---

[11]https://figshare.com/articles/PHEME_dataset_of_rumours_and_non-rumours/4010619.

Work by Tolosi et al. (2016) using feature analysis on rumours across different events found it difficult to distinguish rumours and non-rumours as features change dramatically across events. These findings at the tweet level were then resolved by Zubiaga et al. (2016b) showing that generalisability can be achieved by leveraging context of the events.

McCreadie et al. (2015) studied the feasibility of using a crowdsourcing platform to identify rumours and non-rumours in social media, finding that the annotators achieve high inter-annotator agreement. They also categorised rumours into six different types: Unsubstantiated information, disputed information, misinformation/disinformation, reporting, linked dispute, and opinionated. However, their work was limited to crowdsourced annotation of rumours and non-rumours and they did not study the development of an automated rumour detection system. The dataset from this research is not publicly available.

Yet, other work has been labelled as rumour detection, focusing on determining if information posted in social media was true or false, rather than on early detection of unverified information, and hence we discuss this in Section 9 on veracity classification.

*State of the Art.* The state-of-the-art approach for rumour detection is that presented by Zubiaga et al. (2017), which leverages context from earlier posts associated with a particular event to determine if a tweet constitutes a rumour.

## 7 RUMOUR TRACKING

### 7.1 Definition of the Task and Evaluation

The rumour tracking component is triggered once a rumour is detected and consists of identifying subsequent posts associated with the rumour being monitored. The input is usually a stream of posts, which can be tailored to the rumour in question after filtering for relevant keywords, or it can be broader by including posts related to a bigger event or even an unrestricted stream of posts. The task is generally framed as a binary classification task that consists of determining whether each of the posts is related to the rumour or not. The output will be a labelled version of the stream of posts, where labels define if each post is *related* or *unrelated*.

Traditional evaluation methods for binary classification are used for this task, namely precision, recall, and F1 score, where the positive class is the set of *related* posts.

### 7.2 Datasets

The most widely used dataset for rumour tracking is that by Qazvinian et al. (2011), which includes over 10,000 tweets associated with 5 different rumours, each tweet annotated for relevance toward the rumour as *related* or *unrelated*. *Unrelated* tweets have similar characteristics to those *related*, such as overlapping keywords, and therefore the classification is more challenging.

While not specifically intended for rumour tracking, the dataset produced by Zubiaga et al. (2016) provides over 4,500 tweets categorised by rumour. This dataset is different as it does not include tweets with similar characteristics that are actually *unrelated*. Instead, it provides tweets that are associated with different rumours and tweets that have been grouped by rumour.

### 7.3 Approaches to Rumour Tracking

Research in rumour tracking is scarce in the scientific literature. Despite early work by Qazvinian et al. (2011) performing automated rumour tracking, few studies have subsequently followed their line of research when it comes to determining the relevance of tweets to rumours. Qazvinian et al. (2011) use a manually generated Twitter data set containing 10K tweets to guide a supervised machine learning approach. The authors use different features categorised as "content," "network," and "Twitter specific memes." The content category contains unigrams, bigrams, and their part-of-speech (POS) tags as features. In the network category, the authors look at RTs as a feature. Finally, the Twitter-specific memes include content features inferred from hashtags and URLs. A Bayesian classifier is used as the machine learning approach. The best performance was achieved by using content-based features.

Later work by Hamidian and Diab (2015) also focused on a rumour tracker, using the dataset produced by Qazvinian et al. (2011). They used an approach called Tweet Latent Vector (TLV), which creates a latent vector representative of a tweet to overcome the limited length and context of a tweet. Their approach relies on the Semantic Textual Similarity (STS) model proposed by Guo and Diab (2012), which exploits WordNet (Miller 1995), Wiktionary,[12] and Brown clusters (Brown et al. 1992) to enhance the shortage of semantic meaning of a tweet. This approach outperformed the baseline score established earlier by Qazvinian et al. (2011).

Rumour tracking has not been studied for emerging rumours. The most relevant work to that of tracking newly emerging rumours is that conducted for event detection and tracking in social media (Jaidka et al. 2016). For instance, Sayyadi et al. (2009) describe an event detection and tracking approach based on keyword graphs. They build a graph of keywords to detect communities and subsequently newly emerging events. They then use the set of keywords associated with an event to track new incoming tweets. Similar approaches to event tracking have been introduced by others, such as using a bipartite graph for topical word selection (Long et al. 2011), using text classification techniques to determine whether incoming data is related to a previously identified event or to a new one (Reuter and Cimiano 2012), and using similarity metrics (Tzelepis et al. 2016). However, these approaches have not been directly applied to rumours and hence their applicability needs to be further studied with a suitable rumour dataset.

*State of the Art.* The best approach to rumour tracking is that by Hamidian and Diab (2015) using the tweet latent vector approach. However, more work is still needed on rumour tracking to develop generalisable approaches, especially enabling tracking of newly emerging rumours, which has not yet been studied.

## 8 RUMOUR STANCE CLASSIFICATION

### 8.1 Definition of the Task and Evaluation

The rumour stance classification task consists of determining the type of orientation that each individual post expresses toward the disputed veracity of a rumour. The task is especially interesting in the context of social media, where unverified reports are continually being posted and discussed, both on breaking news stories as they unfold as well as on long-standing rumours. A rumour stance classifier usually takes a set of rumours $D = \{R_1, \ldots, R_n\}$, where each rumour is composed of a collection of posts discussing it. Each rumour has a variably sized set of posts $t_i$ discussing it so that $R_i = \{t_1, \ldots, t_{|R_i|}\}$; the task consists of determining the stance of each of the posts $t_j$ pertaining to a rumour $R_i$. The classification scheme to determine the stance of each post varies across different studies; while early work (Qazvinian et al. 2011) performed two-way classification of $Y = \{supporting, denying\}$, later work performed three-way classification (Lukasik et al. 2015a) involving $Y = \{supporting, denying, querying\}$ as well as four-way classification (Zubiaga et al. 2016a) into $Y = \{supporting, denying, querying, commenting\}$.

The evaluation of the rumour stance classifier is usually based on micro-averaged precision, recall, and F1 scores, as well as accuracy scores. However, as research has progressed into a four-way classification, which generally shows a skewed distribution of labels, evaluation is now also focusing on macro-averaged scores for a fairer evaluation, rewarding the classifiers that perform well across the different labels. More details on the rumour stance classification task can be found in the report of the RumourEval shared task (Derczynski et al. 2017).

---

[12]https://www.wiktionary.org/.

### 8.2 Datasets

Work on stance classification has made use of different datasets, although only two of these datasets are publicly available. One is the PHEME stance dataset,[13] which provides tweet-level annotations of stance (support, deny, query, comment) for tweets associated with nine events. The other publicly available dataset is Ferreira and Vlachos (2016); this dataset does not provide social media data, but it may be leveraged for social media rumour classification as it contains 300 rumoured claims and 2,595 associated news articles, collected and labelled by journalists, along with an estimation of their veracity. Other datasets used in previous work include a dataset with over 10,000 tweets annotated for stance as support, deny, or query by Qazvinian et al. (2011) and the dataset annotated as affirm, deny, neutral, uncodable, or unrelated by Andrews et al. (2016); however, the latter two are not publicly available.

A dataset released for the Fake News Challenge[14] is also annotated for stance (agrees, disagrees, discusses, unrelated). This dataset is, however, made of news articles instead of social media posts.

### 8.3 Approaches to Rumour Stance Classification

Stance classification is well studied in online debates where the aim is to classify the user entries as "for" or "against." Studies in this respect define stance as an overall position held by a person toward an object, idea, or position (Somasundaran and Wiebe 2009; Walker et al. 2012). Unlike stance classification in online debates, the aim of rumour stance classification is to classify user contributions as, e.g., "*supporting,*" "*denying,*" "*querying,*" or "*commenting.*" However, in the literature, the *querying* and *commenting* categories are sometimes omitted or replaced by the "*neutral*" label that encompasses everything that is not *supporting* or *denying*. The rumour stance classification task has attracted many studies over the past few years. All studies follow a supervised approach, and mainly differ in the way they represent a post and how this representation is used to generate a predictive model, i.e., in the features and in the machine learning approaches used to learn predictive models.

One of the pioneering studies in this task was reported by Mendoza et al. (2010). The study involved a human-labelled, non-automated analysis of rumours with established veracity levels to understand the stance that Twitter users express with respect to true and false rumours. The authors looked at 14 rumours, 7 of which turned out to be true and the other 7 were proven false. They manually labelled the tweets associated with those rumours with the stance categories "*affirms*" (supports), "*denies,*" and "*questions.*" They found that over 95% of tweets associated with true rumours were "*affirms,*" whereas only 4% were "*questions,*" and only 0.4% were "*denies.*" This suggested that true rumours are largely supported by other Twitter users. On the other hand, 38% of the tweets associated with false rumours were identified as "*denies*" and 17% as "*questions.*" While false rumours are not denied as often as true rumours are supported, both of these figures suggest that there is indeed a difference in the stances expressed by users toward true and false rumours and that user stances can be indicative of rumour veracity. This study aims to understand the stance categories by manual classification and human analysis, so it does not propose any solution to perform the stance classification task automatically. The first study that tackled the stance classification automatically was reported by Qazvinian et al. (2011). In addition to stance classification, the authors also performed automatic rumour tracking, as we reported in Section 7. The supervised approach developed for the rumour tracking task was also adopted for the stance classification task. In the rumour stance classification task, the tweets were classified as *supporting*, *denying*, *questioning*, or *neutral*. In terms of results, observations similar to the ones obtained for the rumour tracker were reported. The authors reported best results when all features were combined. As with the rumour tracking task, among all features the best performing were those belonging to the content category.

---

[13]https://figshare.com/articles/PHEME_rumour_scheme_dataset_journalism_use_case/2068650.
[14]http://www.fakenewschallenge.org/.

Like Qazvinian et al. (2011), the work by Hamidian and Diab (2015) reported rumour tracking and rumour stance classification by applying supervised machine learning using the dataset created by Qazvinian et al. (2011). However, instead of Bayesian classifiers, the authors used the J48 decision tree implemented within the Weka platform (Hall et al. 2009). The features from Qazvinian et al. (2011) were adopted and extended with time-related information and the hashtag itself as a token instead of the semantic content of the hashtag as used by Qazvinian et al. (2011). In addition to these features, Hamidian and Diab introduced another category: pragmatics. The pragmatic features included named entities, events, sentiment, and emoticons. The evaluation of the performance was cast either as a one-step problem containing a six-way classification task (unrelated to rumour, four classes of stance and not determined) or as a two-step problem containing first a three-way classification task (related to rumour, unrelated to rumour, not determined) and then four-class classification task (stance classification). The highest performance scores were achieved using the two-step approach. The authors also reported that the best performing were the content-based features and the worst performing ones were network and Twitter specific features. In their recent paper, Hamidian and Diab (2016) introduced the TLV approach that is obtained by applying the Semantic Textual Similarity model proposed by Guo and Diab (2012). The authors compared the TLV approach to their own earlier system as well as to original features of Qazvinian et al. (2011) and showed that the TLV approach outperforms both baselines.

Liu et al. (2015) followed the investigations of stances in rumours by Mendoza et al. (2010) and used stance as an additional feature to those reported in related work to tackle the veracity classification problem (see Section 9). For stance classification, the authors adopted the approach of Qazvinian et al. (2011) and compared it with a rule-based method briefly outlined by the authors. They claimed that their rule-based approach performed better than that adopted in previous work, and thus used the rule-based stance classification as an additional component in the veracity problem (see Section 9). The experiments were performed on the dataset reported by Qazvinian et al. (2011). Unfortunately, the authors did not provide a detailed analysis of the performance of the stance classifier.

More recently, Zeng et al. (2016) enriched the feature sets investigated by earlier studies by features determined through Linguistic Inquiry and Word Count (LIWC) dictionaries (Tausczik and Pennebaker 2010). They investigated supervised approaches to stance classification using logistic regression, naïve Bayes, and random forest classification. The authors used their own manually annotated data to classify tweets for stance. However, unlike previous studies, Zeng et al. considered only two classes: affirm and deny. The best results were achieved using a random forest classifier.

Lukasik et al. (2016) investigated Gaussian processes as rumour stance classifiers. For the first time, the authors also used Brown clusters to extract the features for each tweet. The authors used rumour data released by Zubiaga et al. (2016c) and report an accuracy of 67.7%. The authors performed training on $n-1$ rumours and testing on the $n^{th}$ rumour. However, better results were achieved when a small proportion from the in-domain data (data from the $n^{th}$ rumour) was included in the training, leading to an increase in accuracy of around 2%.

Subsequent work has also tackled stance classification for new, unseen rumours. Zubiaga et al. (2016a) moved away from the classification of tweets in isolation, focusing instead on Twitter conversations (Tolmie et al. 2018) initiated by rumours. They looked at tree-structured conversations initiated by a rumour and followed by tweets, responding to it by supporting, denying, querying, or commenting on it. To mine the conversational nature of the data, they used CRFs as a sequential classifier in two different settings: linear-chain CRFs and tree CRFs. Their objective with CRFs was to exploit the discursive nature of the argumentation produced collaboratively by users. Their experiments on eight different datasets of rumours spread during breaking news showed that the discursive characteristics of conversations can indeed be exploited with a sequential classifier to improve on the performance that equivalent, non-sequential classifiers can achieve.

Rumour stance classification for tree structured conversations has also been studied in the RumourEval shared task at SemEval 2017 (Derczynski et al. 2017). The subtask A consisted of stance classification of individual tweets discussing a rumour within a conversational thread as one of *support*, *deny*, *query*, or *comment*. Eight participants submitted results to this task. Most of the systems viewed this task as a four-way single tweet classification task, with the exception of the best performing system by Kochkina et al. (2017), as well as the systems by Wang et al. (2017) and Singh et al. (2017). The winning system addressed the task as a sequential classification problem, where the stance of each tweet takes into consideration the features and labels of the previous tweets. The system by Singh et al. (2017) takes source and reply tweets as input pairs, whereas Wang et al. (2017) addressed class imbalance by decomposing the problem into a two-step classification task, first distinguishing between comments and non-comments, to then classify non-comment tweets as one of support, deny, or query. Half of the systems employed ensemble classifiers, where classification was obtained through majority voting (Wang et al. 2017; García Lozano et al. 2017; Bahuleyan and Vechtomova 2017; Srivastava et al. 2017). In some cases the ensembles were hybrid, consisting both of machine learning classifiers and manually created rules, with differential weighting of classifiers for different class labels (Wang et al. 2017; García Lozano et al. 2017; Srivastava et al. 2017). Three systems used deep learning, with Kochkina et al. (2017) employing Long/Short-Term Memory Networks (LSTMs) for sequential classification; Chen et al. (2017), using convolutional neural networks (CNN) for obtaining the representation of each tweet, assigned a probability for a class by a softmax classifier; and García Lozano et al. (2017) using CNN as one of the classifiers in their hybrid conglomeration. The remaining two systems by Enayet and El-Beltagy (2017) and Singh et al. (2017) used support vector machines with a linear and polynomical kernel, respectively. Half of the systems invested in elaborate feature engineering, including cue words and expressions denoting belief, knowledge, doubt, and denial (Bahuleyan and Vechtomova 2017) as well as tweet domain features, including metadata about users, hashtags, and event-specific keywords (Wang et al. 2017; Bahuleyan and Vechtomova 2017; Singh et al. 2017; Enayet and El-Beltagy 2017). The systems with the least-elaborate features were Chen et al. (2017) and García Lozano et al. (2017) for CNNs (word embeddings), Srivastava et al. (2017) (sparse word vectors as input to logistic regression) and Kochkina et al. (2017) (average word vectors, punctuation, similarity between word vectors in current tweet, source tweet, and previous tweet, presence of negation, picture, URL). Five out of the eight systems used pre-trained word embeddings, mostly Google News word2vec embeddings,[15] whereas Wang et al. (2017) used four different types of embeddings.

Other related studies have looked into stance classification not directly applicable to rumour stance classification. While Zhao et al. (2015) did not study stance classification per se, they developed an approach to look for querying tweets, which is one of the reaction types considered in stance classification. However, the other stance types were not considered and querying tweets were found by matching with manually defined regular expressions, which may not be directly applicable to other stance types. While not focused on rumours, classification of stance toward a target on Twitter was addressed in SemEval-2016 Task 6 (Mohammad et al. 2016). Task A had to determine the stance of tweets toward five targets as "favor," "against," or "none." Task B tested stance detection toward an unlabelled target, which required a weakly supervised or unsupervised approach.

---

[15]https://github.com/mmihaltz/word2vec-GoogleNews-vectors.

Researchers have also studied the identification of agreement and disagreement in online conversations. To classify agreement between question-answer (Q-A) message pairs in fora, Abbott et al. (2011) usednaïve Bayes as the classifier and Rosenthal and McKeown (2015) used a logistic regression classifier. A sequential classifier like CRF has also been used to detect agreement and disagreement between speakers in broadcast debates (Wang et al. 2011). It is also worthwhile emphasising that stance classification is different to agreement/disagreement detection, given that, in stance classification, the orientation of a user toward a rumour has to be determined. In contrast, in agreement/disagreement detection, it is necessary to determine if a pair of posts share the same view. In stance classification, a user might agree with another user who is denying a rumour and hence they are denying the rumour as well, irrespective of the pairwise agreement.

*State of the Art.* The majority of the work on stance classification performed experiments on different datasets and evaluation methods, so that it is hard to pick the state-of-the-art method among them. However, some conclusions can be drawn on the RumourEval stance shared task, where the approach by Kochkina et al. (2017) was judged as the best performing method for stance classification for Twitter rumours. More recently, Aker et al. (2017) reported state-of-the-art results on the same dataset. Their approach is simple and profits from a novel set of automatically identifiable problem-specific features, which significantly boosted classifier accuracy and achieved better results than those reported by Kochkina et al. (2017).

## 9 RUMOUR VERACITY CLASSIFICATION

### 9.1 Definition of the Task and Evaluation

The veracity classification task aims to determine whether a given rumour can be confirmed as true, debunked as false, or its truth value is still to be resolved. Given a set of posts associated with a rumour and, optionally, additional sources related to the rumour, the task consists of assigning one of the following labels to the rumour, $Y \in \{true, false, unverified\}$. Some work has limited the classification to a binary task of determining if a rumour is true or false; however, it is likely that veracity value will remain uncertain for some rumours. Optionally, the classifier can also output, along with the veracity label, the confidence with which the label has been assigned, usually ranging from 0 to 1.

The outcome of the veracity classification task is usually evaluated either using the accuracy measure that computes the ratio of correct classifications, or using a combination of precision, recall, and F1 score for the three categories.

### 9.2 Datasets

The dataset produced for RumourEval 2017 (Derczynski et al. 2017), a shared task that took place at SemEval 2017, includes over 300 rumours annotated for veracity as one of *true, false,* or *unverified*. Another dataset suitable for veracity classification is that released by Kwon et al. (2017), which includes 51 true rumours and 60 false rumours. Each rumour includes a stream of tweets associated with it.

Other datasets, such as that by Qazvinian et al. (2011), are not suitable for veracity classification, as all the rumours are false.

## 9.3 Approaches to Rumour Veracity Classification

The vast majority of research dealing with social media rumours has focused on veracity classification, which is the crucial and ultimate goal of determining the truth value of a circulating rumour. This work generally assumes that rumours have already been identified or input by a human. Therefore, most of the previous work skips the preceding components of a rumour classification system, especially the rumour detection component that identifies candidate rumours to be input to the veracity classification system.

Work by Castillo et al. (2011) initiated research in this direction of determining the veracity of social media content, although the authors did not directly tackle the veracity of rumours but rather their credibility perceptions, i.e., determination of the believability or authority of its source (Zhang et al. 2015). However, others report that veracity is related to authority (American Public Health Association et al. 2001; Oh et al. 2013) and hence Castillo et al.'s work has been considered as a reference by many in subsequent work on veracity classification.

To study credibility perceptions, Castillo et al. (2011) distinguished two types of microblog posts: NEWS, which reports an event or fact that can be of interest to others, and CHAT, which is a message that is purely based on personal/subjective opinions and/or conversations among friends. Microblogs categorised as NEWS were analysed for rumour credibility. Decision trees based on J48 were used to train classifiers and these were used to classify microblogs into NEWS and CHAT categories. The microblogs in the NEWS category were further analysed for credibility—for this task, the authors reported that they used other various machine learning approaches, such as Bayesian networks and SVM classifiers, but noted that decision trees based on J48 were superior. In the experiments, microblogs collected from Twitter were used. These were manually annotated using Mechanical Turk. The authors used four categories of features: message-based, user-based, topic-based, and propagation-based features. The message-based features considered characteristics of messages such as the length of a message, whether the message contained exclamation/question marks, number of positive/negative sentiment words, whether the message contained a hashtag, and whether it was an RT. User-based features entailed information such as registration age, number of followers, number of followees, and the number of tweets the user had authored in the past. Topic-based features aggregated information from the previous two feature types, such as the fraction of tweets that contained URLs, the fraction of tweets with hashtags, and the like. Finally, the propagation-based features considered characteristics related to the messaging tree, such as depth of the RT tree or the number of initial tweets on a topic. The authors reported highly accurate results on both tasks. Despite the fact that this approach was designed for the classification of credibility perceptions, the features utilised in this work have subsequently also been exploited for veracity classification.

Most research, however, has dealt with veracity classification. Kwon et al. (2013) proposed a new set of feature categories: temporal, structural, and linguistic. The temporal features aimed to capture how rumours spread over time. The structural features model the connectivity between users who posted about the rumour. Finally, the linguistic features were obtained through the LIWC dictionaries (Tausczik and Pennebaker 2010). As a baseline classifier, features proposed by Castillo et al. (2011) were adopted. Using random forest and logistic regression, the authors performed feature selection to find the most significant ones. Using these features and three different classifiers (decision tree, random forest, and SVM) they then performed rumour veracity classification. The results showed that for both selection of significant features and subsequent classification, a random forest classifier performed best. The best results using the baseline features adopted from Castillo et al. (2011) were obtained using an SVM. The authors also showed that a combination of significant features identified by random forest and baseline features led to performance degradation. More recently, Kwon et al. (2017) analysed feature stability over time and reported that structural and temporal features distinguished true from false rumours over a long-term window. However, the authors also reported that these were not available in the early stages of rumour propagation but only later on. In contrast, user and linguistic features are an alternative when the task is to determine rumour veracity as early as possible.

Yang et al. (2012) tackled the veracity of microblogs on the Chinese microblogging platform Sina Weibo. The authors adopted features from earlier studies discussed above and extended them with two more features: client-based and location-based features. The client-based features included information about the software that was used to perform the messaging. The location-based features included information relating to whether the message was sent from within the same country where the event happened or not. The authors reported that adding these two features on top of earlier reported features led to a substantial boost of accuracy. For instance, adding the two features on top of the propagation-based features reported by Castillo et al. (2011) led to an increase of 6.3% in accuracy (from 72.3% to 78.6%). However, the authors did not combine all the features and reported results on them. For the classification task, the authors use Support Vector Machines (SVM) with the Radial Basis Function (RBF) kernel (Buhmann 2003). Another study that tackled rumours in Sina Weibo is reported by Yang et al. (2015). Unlike the others, the authors made use of the reviews or comments attached to the source tweet. A number of features discussed so far are used, but they also incorporated network features (creating a social network based on the comment providers) derived from comments to perform the rumour veracity classification task. It was shown that when the network feature was added to the traditional features, the results improved substantially.

Liu et al. (2015) used approaches reported by Yang et al. (2012) and Castillo et al. (2011) as baseline systems and compared them against their proposed approach that make use of so-called verification features. These features were determined based on insights from journalists and included source credibility, source identification, source diversity, source and witness location, event propagation, and belief identification. In belief identification, results of rumour stance classification were used as features. The authors showed that the proposed approach outperformed the two baselines. They also showed that, when adding the belief identification to the other features, the results were significantly better than when not adding them to the feature set. The experiments were performed on the author's own dataset using SVM classification. The authors also reported having investigated random forest and decision trees but SVM gave the best results, although no further details were provided on this comparison.

Ma et al. (2015) proposed modelling features over time. The authors adopted features from earlier studies, as well as machine learning approaches used in those studies (J48, SVM with the RBF kernel). Experiments were performed on datasets from both Twitter and Sina Weibo. Ma et al. used SVM with linear kernel and reported that this linear SVM, combined with modelling the features over time, led to the best performance. For Sina Weibo, the authors collected their own dataset and ran existing approaches on them. The best results were obtained using the proposed method reaching an accuracy of 84.6%. Decision trees (J48), SVM, and random forests achieved around 3–7% less accuracy. Wu et al. (2015) extracted features from message propagation trees. Three categories of features were considered: message-based, user-based, and report-based. Two methods reported by earlier studies ((Castillo et al. 2011) and Yang et al. (2012)) were adopted for the evaluation. The machine learning approach chosen by the authors was an SVM with a hybrid kernel technique consisting of random walk kernel (Borgwardt et al. 2005) and an RBF kernel. The results reported were in favour of the proposed hybrid approach. In the experiments, the authors used rumours with at least 100 reposts. This posed the question as to how well the proposed approach would perform when it is applied to newly emerging rumours where there are only few posts available. The idea of message propagation was also investigated by Wang and Terano (2015) in combination with pattern matching. In addition, the information inferred from a stance classifier was integrated in the classification process. They proposed social graphs to model the interaction between users and so identify influential rumour spreaders. The graph entailed information about familiarity measured by the number of contacts such as RTs, replies, and comments between two users, activeness measured by the number of days a user has sent out messages, similarity measured by gender and location, similarity between two users, and trustworthiness measured by whether the user is verified or not. These four factors were merged in a linear model and hence the model was used to weight the link between two users in the social graph. Using a new proposed metric, influential spreaders were used to determine rumours.

Vosoughi (2015) tackled veracity classification tasks using three categories of features (linguistic, user oriented, and temporal propagation related) and speech recognition inspired machine learning approaches, such as dynamic time wrapping (DTW) and hidden Markov models (HMMs). Evaluations were performed on Twitter datasets gathered by the author. The results showed that HMMs were superior to DTWs. The authors also reported that the best performing features were those in the temporal propagation category. The linguistic and user-oriented features did not contribute as much as those in the temporal propagation category. It is also worthwhile noting that researchers, such as Wu et al. (2015), worked with rumours that had a good volume of tweets, in this case at least 1,000.

Giasemidis et al. (2016) reported experiments run on 100 million tweets associated with 72 different rumours. Features and machine learning classifiers used in previous work were adopted in this case. The authors reported very good results using decision trees. Chen et al. (2016) approached rumour veracity classification from a different angle. The authors treated it as an anomaly detection problem where false rumours are regarded as anomalies. Several features related to the content, crowd opinion, and post propagation were used, along with a factor analysis. Euclidean distance and cosine similarity were proposed to describe the deviation degree, and posts with high deviation degree were marked as rumours. Comparisons were made with respect to well-known clustering approaches, such as K-means, and the reported results showed significantly improved performance.

Chang et al. (2016) put the emphasis on the characteristics of users who post the rumours to determine the veracity. The authors focused on tweets discussing news. Such tweets were first clustered using simple heuristics, e.g., all posts linking to the same news article were grouped together. Based on rules and heuristics, "extreme users" were determined—users who matched the heuristics, such as number of followers and so on, of users that were likely to post false rumours. If a cluster of posts contained a number of extreme users exceeding a predefined threshold, then it was marked as a false rumour cluster.

Chua and Banerjee (2016) published an analysis of various features on the tweet veracity classification task. The authors analysed six categories of features: comprehensibility, sentiment, time-orientation, quantitative details, writing style, and topic. Rumours gathered by the authors were used along with the binomial logistic regression to tackle the task in a supervised fashion. Unlike previous studies, Chua and Banerjee (2016) only reported features that are significantly important, rather than an indication of the overall performance of the classifier. These features were: negation words (comprehensibility category), past, present, future POS in the tweets (time-orientation category), discrepancy, sweat and exclusion features (writing style category), and, finally, home, leisure, religion, and sex topic features (topic category). In a similar vein, Ma et al. (2017) investigated the performance difference between bag-of-words (BoW) and word-embedding representation of post contents, and concluded that the BoW representation was superior to the embedding variant.

Zhang et al. (2015) investigated rumour veracity classification within the health domain. Using data obtained from liuyanbaike.com (a Chinese rumour-debunking platform), the authors investigated the correlation between features and veracity of rumours (true or false) based on logistic regression. They reported that features like mention of numbers, the source the rumour originated from, and hyperlinks, positively correlated with true rumours and rumours containing some wishes, were positively correlated with false rumours. If images were included in the rumours, then those were negatively correlated with true rumours. Finally, the authors reported that the foreign source feature (whether a foreign source was used to support the rumour or not) was not correlated at all with rumour veracity.

Qin et al. (2016) aimed to detect new rumours and proposed two new feature categories to achieve this. In the first category, posts containing new pieces of information that are unconfirmed with respect to some news event were considered as rumours. In the second feature category, later posts that repeated the same information as the earlier ones marked as rumour were also considered as rumours. The authors reported significantly better results than baseline approaches when the aim was to determine veracity of rumours early on (in their particular case, in less than 12 hours).

Unlike the previous studies, Tong et al. (2017) aimed at blocking rumours rather than detecting them or marking tweets as true or false. Motivated by the fact that later corrections are not as effective, the authors argued that the first post seen by a user is influential for their future opinion and thus it is important to show users rumours only once they are confirmed to be true. Based on this, they proposed a reverse-tuple-based randomised algorithm to block rumours. The algorithm aimed at producing positive seeds to be shown to users first.

Rumour veracity classification has also been studied in the RumourEval shared task at SemEval 2017 (Derczynski et al. 2017). Subtask B consisted of determining if each of the rumours in the dataset were true, false, or remained unverified. It considered two different settings, one *closed* where participants could not make use of external knowledge bases, and another *open* where use of external resources was allowed. Five participants submitted results to this subtask. Participants viewed the task either as a three-way (Enayet and El-Beltagy 2017; Wang et al. 2017; Singh et al. 2017) or two-way (Chen et al. 2017; Srivastava et al. 2017) single tweet classification task. The methods used mostly the same features and classifiers as used in subtask A (see Section 8), although some added features more specific to the distribution of stance labels in the tweets replying to the source tweet (for example, the best performing system in this task (Enayet and El-Beltagy 2017) considered the percentage of reply tweets classified as either support, deny, or query).

*State of the Art.* The selection of the best-performing system is not easy, due to the use of different datasets and evaluation methods. However, we can again consider the RumourEval shared task to determine the best system for the veracity verification task. As discussed above, the approach reported by Enayet and El-Beltagy (2017), which aggregates the stance of individual tweets to determine the veracity of rumours as a three-way classification problem, is currently the best performing system.

## 10 APPLICATIONS

There have been numerous efforts by both industry and the scientific community to deal with social media rumour detection and verification, ranging from ongoing research projects to fully fledged applications. The following are some notable examples:

—PHEME[16] (Derczynski and Bontcheva 2014) is a 3-year research project funded by the European Commission, which ran from 2014-2017, studying natural language processing techniques for dealing with rumour detection and resolution. Publications produced as part of this project include rumour detection (Zubiaga et al. 2016b), stance classification (Lukasik et al. 2015a, 2016; Zubiaga et al. 2016a), contradiction detection (Lendvai et al. 2016a; Lendvai and Reichel 2016), ontological modelling of rumours (Declerck et al. 2015), visualisation (Lendvai et al. 2016b), analysis of social media rumours (Zubiaga et al. 2016c), and studies of journalistic practices of the use of UGC (Tolmie et al. 2017).

---
[16]http://www.pheme.eu/.

—Emergent[17] is a data-driven, real-time, web-based rumour tracker. The system automatically tracks social media mentions of URLs' associated rumours; however, the identification of rumours and selection of URLs associated with those requires human input and has not been automated. It is part of a research project led by Craig Silverman, partnering with the Tow Center for Digital Journalism at Columbia University, which focuses on how unverified information and rumour are reported in the media. The outcome of this project was published in a report on best practices for debunking misinformation (Silverman 2015a).

—RumorLens[18] (Resnick et al. 2014) is a 1-year research project that ran in 2014, funded by Google. It focused on building a tool to aid journalists in finding posts that spread or correct a particular rumour on Twitter, by exploring the size of the audiences that those posts have reached. More details on the rumour detection system developed in this project were published in Zhao et al. (2015).

—TwitterTrails[19] (Finn et al. 2014) is a project in the Social Informatics Lab at Wellesley College. Twitter Trails is an interactive, web-based tool that allows users to investigate the origin and propagation characteristics of a rumour and its refutation, if any, on Twitter. Visualisations of burst activity, propagation timeline, RT, and co-retweeted networks help its users trace the spread of a story. It collects relevant tweets and automatically answers several important questions regarding a rumour: its originator, burst characteristics, propagators, and main actors according to the audience. In addition, it computes and reports the rumour's level of visibility and, as an example of the power of crowdsourcing, the audience's skepticism toward it, which correlates with the rumour's credibility. The project has produced a number of publications (cf. Finn et al. (2015) and Metaxas et al. (2015)) exploring and characterising the diffusion of rumours.

—RumourFlow (Dang et al. 2016b) is a framework that designs, adopts and implements multiple visualisations and modelling tools that can be integrated to reveal rumour contents and participant activity, both within a rumour and across different rumours. The approach supports analysts in drawing hypotheses regarding rumour propagation.

—COSMIC was a 3-year research project funded by the European Commission that studied the contribution of social media to crisis management. As part of the project, the adverse use and reliability of social media was studied, including the impact of rumours (Scifo and Baruh 2015).

—SUPER was a 3-year research project funded by the European Commission that studied the use of social sensors for security assessments and proactive emergencies management, in part dealing with crowdsourced annotation of rumours (McCreadie et al. 2015).

—Hoaxy[20] (Shao et al. 2016) is a platform for the collection, detection and analysis of online misinformation and its related fact-checking efforts.

—REVEAL[21] was a 3-year project (2013–2016) funded by the European Commission. It was concerned with verification of social media content from a journalistic and enterprise perspective, especially focusing on image verification. The project produced a number of publications on journalistic verification practices concerning social media (Brandtzaeg et al. 2016), social media verification approaches (Andreadou et al. 2015), and approaches to track down the location of social media users (Middleton and Krivcovs 2016).

---

[17]http://www.emergent.info/.
[18]https://www.si.umich.edu/research/research-projects/rumorlens.
[19]http://twittertrails.com/.
[20]http://hoaxy.iuni.iu.edu/.
[21]http://revealproject.eu/.

—InVID[22] (In Video Veritas) is a Horizon 2020 project, funded by the European Commission (2017-2020), which will build a platform providing services to detect, authenticate, and check the reliability and accuracy of newsworthy video files and video content spread via social media.

—CrossCheck[23] is a collaborative verification project led by First Draft and Google News Lab, in collaboration with a number of newsrooms in France, to fight misinformation, with an initial focus on the French presidential election.

—Décodex[24] is an online database by the French news organisation Le Monde, which enables checking the reliability of news sites.

—Check[25] is a verification platform that offers newsrooms the possibility of verifying breaking news content online. The platform is not yet openly available, but there is a form to register interest.

—ClaimBuster[26] is a project aiming to perform live fact-checking. The demo application shows check-worthy claims identified by the system for the 2016 U.S. election and it allows the user to input their own text to find factual claims. Details of the project have been published in Hassan et al. (2015).

—Una Hakika[27] is a Kenyan project dealing with misinformation and disinformation. It offers a search engine to look for rumours, as well as an API for data collection. It is manually updated with new stories.

—Seriously Rapid Source Review[28] (Diakopoulos et al. 2012) is a system that incorporates a number of advanced aggregations, computations, and cues that can be helpful to journalists for finding and assessing sources in Twitter around breaking news events, such as finding eyewitnesses to events, which can be helpful to either confirm or debunk rumours.

—TweetCred[29] (Gupta and Kumaraguru 2012; Gupta et al. 2014) is a real-time, web-based system to assess the credibility of content on Twitter. While the system does not determine the veracity of stories, it provides a credibility rating between 1 to 7 for each tweet in the Twitter timeline.

## 11 DISCUSSION: SUMMARY AND FUTURE RESEARCH DIRECTIONS

Research on the development of rumour detection and verification tools has become increasingly popular as social media penetration has increased, enabling both ordinary users and professional practitioners to gather news and facts in a real-time fashion, but with the problematic side effect of the diffusion of information of unverified nature. This survey article has summarised studies reported in the scientific literature toward the development of rumour classification systems, defining and characterising social media rumours, and has described the different approaches to the development of their four main components: (1) rumour detection, (2) rumour tracking,

---

[22] http://www.invid-project.eu/.
[23] https://firstdraftnews.com/crosscheck-launches/.
[24] http://www.lemonde.fr/verification/.
[25] https://meedan.com/en/check/.
[26] http://idir-server2.uta.edu/claimbuster.
[27] http://www.unahakika.org/.
[28] http://www.nickdiakopoulos.com/2012/01/24/finding-news-sources-in-social-media/.
[29] http://twitdigest.iiitd.edu.in/TweetCred/.

(3) rumour stance classification, and (4) rumour veracity classification. In so doing, the survey provides a guide to the state of the art in the development of these components. The survey has focused particularly on the classification of rumours circulating in social media. Most of the general aspects, such as rumour definition and the classification architecture, are generalisable to genres such as news articles. However, the specific approaches described for each of the four components are usually designed for social media and are not necessarily directly applicable to other genres. In what follows, we review the progress achieved so far, the shortcomings of existing systems, outline suggestions for future research, and comment on the applicability and generalisability of rumour classification systems to other kinds of misleading information that also spread in social media.

Research in detection and resolution of rumours has progressed substantially since the proliferation of social media as a platform for information and news gathering. A range of studies have taken very different approaches to understanding and characterising social rumours, and this diversity helps to shed light on the future development of rumour classification systems. Research has been conducted in all four of the components that comprise a rumour classification system, although most have focused on the two last components of the pipeline, namely rumour stance classification and veracity classification. Despite substantial progress in the research field, as shown in this survey, we also show that this is still an open research problem that needs further study. We examine the main open research challenges in the next section.

### 11.1 Open Challenges and Future Research Directions

In recent years, research in rumour classification has largely focused on the later stages of the pipeline, namely rumour stance classification and veracity classification. These are crucial stages; however, they cannot be used without performing the preceding tasks of detecting rumours and tracking posts associated with those rumours. The latter has generally been skipped in previous work, either leaving the development of those components for future work or assuming that rumours and associated posts are input by a human. Aiming to alleviate these initial tasks by avoiding relying entirely on the human-in-the-loop, we argue that future research should focus on rumour detection and tracking. Further research in this direction would then enable development of entirely automated rumour classification systems.

Research in rumour detection should start by testing state-of-the-art event detection techniques in the specific context of rumours. Beyond what event detection systems do, a rumour detection system needs to determine if a detected event constitutes a rumour or not. Determining if an individual social media post reports a rumour or not is challenging if only its content is used. Recent research shows that the use of context (Zubiaga et al. 2017) and interactions (Zhao et al. 2015) can be of help, which are directions that are worth exploring in more detail.

Research into rumour tracking systems is limited and researchers often assume that the keywords used to collect posts associated with rumours are known *a priori*. One evident issue with social media is the use of inconsistent vocabulary across users, e.g., users may indistinctly use *killing* or *shooting* to refer to the same event. Research in expanding data collection is still in its infancy and the use of query expansion approaches through techniques, such as pseudo-relevance feedback, are yet to be explored in detail, further to preliminary research showing its potential (Fresno et al. 2015).

An important limitation toward the development of rumour classification systems has been the lack of publicly available datasets. Along with the recently published datasets we have listed in this survey, we encourage researchers to release their own datasets so as to enable further research over different datasets and so enable the scientific community to compare their approaches with one another.

While many have attempted to automatically determine the veracity value of a rumour, a system that simply outputs the final decision on veracity may not always be sufficient, given that the classifier will inevitably make errors. To make the output of a veracity classifier more reliable, we argue that the system needs to provide a richer output that also includes the reason for the decision (Procter et al. 2013b). A veracity classifier that outputs not only the automatically determined veracity score, but also links to sources where this decision can be corroborated, will be more robust in that it will enable the user to assess the reliability of the classifier's decision and—- if found wanting—to ignore it. The output of a veracity classifier can be enriched, for instance, by using the output of the stance classifier to choose a few supporting and opposing views that can be presented to the user as a summary. Given that achieving a perfectly accurate veracity classifier is an unlikely goal, we argue that research in this direction should focus especially on finding information sources that facilitate the end user to make their own judgement of rumour veracity.

Another caveat of existing veracity classification systems is that they have focused on determining veracity regardless of rumours being resolved. Where rumours have not yet been resolved, the veracity classification task then becomes a prediction task, which may not be reliable for an end user given the lack of evidence to support the system's decision. As rumours have an unverified status in which determining veracity is hard or requires involvement of authoritative sources, future research should look into temporality of rumour veracity determination, potentially attempting to determine veracity soon after evidence can be found.

When it comes to stance classification, recent work has shown the effectiveness of leveraging context in social media streams and conversations to develop state-of-the-art classifiers for the stance of individual posts. Research in this direction is, however, still in its infancy and more research is still needed to best exploit this context for maximising the performance of stance classifiers. Research in rumour classification has largely relied on the content of social media posts, while further information extracted from user metadata and interactions may be of help to boost the performance of classifiers.

### 11.2 Rumours, Hoaxes, Misinformation, Disinformation, Fake News

In this survey, we have covered previous efforts toward the development of a rumour classification system that can detect, and resolve the veracity of, rumours. As defined in Section 1.1, rumours refer to pieces of information that start of as unverified statements. A rumour's veracity value is unverifiable in the early stages, while being subsequently resolved as true or false in a relatively short period of time, or it can also remain unverified for a long time. A number of similar terms are also used in related literature, which have distinct characteristics but also commonalities with rumours.

The term misinformation is used to refer to circulating information that is accidentally false as a consequence of an honest mistake, while disinformation refers to information that is deliberately false (Hernon 1995). Rumours can fall in either of these two categories, depending on the intent of the source; however, the main difference is that rumours are not necessarily false, but may turn out to be true. A rumour that is eventually debunked can then be categorised into misinformation or disinformation depending on the intent of the source.

Unlike rumours, hoaxes and fake news are, by definition, always false and can be seen as specific types of disinformation. While it is usually used to refer to any fabricated falsehood indistinctly, a hoax is more rigorously defined as a false story used to masquerade the truth, originating from the verb *hocus*, meaning "to cheat" (Nares 1822). Fake news is a specific type of hoax, usually spread through news outlets that are intended to gain politically or financially (Hunt 2016). However, terms like *fake news* are widely being used to refer to different types of inaccurate information, while not necessarily adhering to any specific type of misinformation. As Wardle (2017) suggested, the term fake news is being used to refer to seven types of misinformation: false connection, false context, manipulated content, satire or parody, misleading content, imposter content, and fabricated content.

The approaches described in this survey article are designed to tackle the problem of rumour. Further research is needed to study their applicability to other phenomena, such as hoaxes and fake news. However, we believe that some of the underlying commonalities between rumours, hoaxes, and fake news suggest that rumour research has an important contribution to make the new challenges posed by these more recent phenomena.

### 11.3 Further Reading

For more discussion on the issues we cover in this survey article, we recommend the special issue on trust and veracity of information on social media of the ACM TOIS journal (Papadopoulos et al. 2016; Rijke 2016), Full Fact's report on *The State of Automated Factchecking* (Babakar and Moy 2016), reports and discussion of rumours on Snopes,[30] and Craig Silverman's books on rumours and journalistic verification practices (Silverman 2013, 2015a, 2015b). We also recommend keeping track of ongoing initiatives by the Knight Center for Journalism in the Americas[31] and the European Journalism Centre,[32] http://ejc.net/., as well as signing up for relevant newsletters such as Craig Silverman's on online rumours, fake news, and misinformation,[33] and Poynter and American Press Institute's *The Week in Fact-Checking*.[34]

**REFERENCES**

Rob Abbott, Marilyn Walker, Pranav Anand, Jean E. Fox Tree, Robeson Bowmani, and Joseph King. 2011. How can you say such things?!?: Recognizing disagreement in informal political argument. In *Proceedings of the Workshop on Languages in Social Media*. 2–11.

Sheetal D. Agarwal, W. Lance Bennett, Courtney N. Johnson, and Shawn Walker. 2014. A model of crowd enabled organization: Theory and methods for understanding the role of twitter in the occupy protests. *Int. J. Commun.* 8 (2014), 27.

Ahmet Aker, Leon Derczynski, and Kalina Bontcheva. 2017. Simple open stance classification for rumour analysis. *arXiv:1708.05286* (2017).

Gordon W. Allport and Leo Postman. 1946. An analysis of rumor. *Publ. Opin. Q.* 10, 4 (1946), 501–517.

Gordon W. Allport and Leo Postman. 1947. *The Psychology of Rumor*. Russell & Russell.

American Public Health Association et al. 2001. Criteria for assessing the quality of health information on the internet. *American Journal of Public Health* 91, 3 (2001), 513.

Katerina Andreadou, Symeon Papadopoulos, Lazaros Apostolidis, Anastasia Krithara, and Yiannis Kompatsiaris. 2015. Media REVEALr: A social multimedia monitoring and intelligence system for web multimedia verification. In *Pacific-Asia Workshop on Intelligence and Security Informatics*. 1–20.

Cynthia Andrews, Elodie Fichet, Yuwei Ding, Emma S. Spiro, and Kate Starbird. 2016. Keeping up with the tweet-dashians: The impact of "official" accounts on online rumoring. In *Proceedings of the 19th ACM Conference on Computer-Supported Cooperative Work & Social Computing*. ACM, 452–465.

Nick Anstead and Ben O'Loughlin. 2015. Social media analysis and public opinion: The 2010 UK general election. *J. Comput.-Med. Commun.* 20, 2 (2015), 204–220.

Ahmer Arif, Kelley Shanahan, Fang-Ju Chou, Yoanna Dosouto, Kate Starbird, and Emma S. Spiro. 2016. How information snowballs: Exploring the role of exposure in online rumor propagation. In *Proceedings of the ACM Conference on Computer-Supported Cooperative Work & Social Computing*. ACM, 466–477.

Pablo D. Azar and Andrew W. Lo. 2016. The wisdom of twitter crowds: Predicting stock market reactions to fomc meetings via twitter feeds. *The Journal of Portfolio Management* 42, 5 (2016), 123–134.

Mevan Babakar and Will Moy. 2016. The state of automated factchecking. *Full Fact* (2016).

---

[30]http://www.snopes.com/.
[31]https://knightcenter.utexas.edu/.
[32]http://ejc.net/.
[33]http://us2.campaign-archive1.com/?u=657b595bbd3c63e045787f019&id=2208e04aa6.
[34]http://us9.campaign-archive1.com/?u=79fa45ed20ff84851c3b9cd63&id=02624abd8b.

Hareesh Bahuleyan and Olga Vechtomova. 2017. UWaterloo at SemEval-2017 Task 8: Detecting stance towards rumours with topic independent features. In *Proceedings of SemEval*. ACL.

Prashant Bordia. 1996. Studying verbal interaction on the internet: The case of rumor transmission research. *Behav. Res. Methods, Instrum. Comput.* 28, 2 (1996), 149–151.

Karsten M. Borgwardt, Cheng Soon Ong, Stefan Schönauer, S. V. N. Vishwanathan, Alex J. Smola, and Hans-Peter Kriegel. 2005. Protein function prediction via graph kernels. *Bioinformatics* 21 (2005), i47–i56.

Petter Bae Brandtzaeg, Marika Lüders, Jochen Spangenberg, Linda Rath-Wiggins, and Asbjørn Følstad. 2016. Emerging journalistic verification practices concerning social media. *Journal. Pract.* 10, 3 (2016), 323–342.

Peter F. Brown, Peter V. Desouza, Robert L. Mercer, Vincent J. Della Pietra, and Jenifer C. Lai. 1992. Class-based n-gram models of natural language. *Comput. Ling.* 18, 4 (1992), 467–479.

Martin D. Buhmann. 2003. Radial basis functions: Theory and implementations. *Camb. Monog. Appl. Comput. Math.* 12 (2003), 147–165.

Guoyong Cai, Hao Wu, and Rui Lv. 2014. Rumors detection in chinese via crowd responses. In *Proceedings of the 2014 IEEE/ACM International Conference on Advances in Social Networks Analysis and Mining (ASONAM'14)*. IEEE, 912–917.

Carlos Castillo. 2016. *Big Crisis Data: Social Media in Disasters and Time-Critical Situations*. Cambridge University Press.

Carlos Castillo, Marcelo Mendoza, and Barbara Poblete. 2011. Information credibility on twitter. In *Proceedings of the 20th International Conference on World Wide Web*. ACM, 675–684.

Carlos Castillo, Marcelo Mendoza, and Barbara Poblete. 2013. Predicting information credibility in time-sensitive social media. *Internet Res.* 23, 5 (2013), 560–588.

Cheng Chang, Yihong Zhang, Claudia Szabo, and Quan Z. Sheng. 2016. Extreme user and political rumor detection on twitter. In *Advanced Data Mining and Applications: 12th International Conference, ADMA 2016, Gold Coast, QLD, Australia, December 12-15, 2016, Proceedings*. Springer, 751–763.

Hailiang Chen, Prabuddha De, Yu Jeffrey Hu, and Byoung-Hyoun Hwang. 2014. Wisdom of crowds: The value of stock opinions transmitted through social media. *Rev. Financial Stud.* 27, 5 (2014), 1367–1403.

Weiling Chen, Chai Kiat Yeo, Chiew Tong Lau, and Bu Sung Lee. 2016. Behavior deviation: An anomaly detection view of rumor preemption. In *Proceedings of the 2016 IEEE 7th Annual Information Technology, Electronics and Mobile Communication Conference (IEMCON'16)*. IEEE, 1–7.

Yi-Chin Chen, Zhao-Yand Liu, and Hung-Yu Kao. 2017. IKM at SemEval-2017 Task 8: Convolutional neural networks for stance detection and rumor verification. In *Proceedings of SemEval*. ACL.

Jun-Jun Cheng, Yun Liu, Bo Shen, and Wei-Guo Yuan. 2013. An epidemic model of rumor diffusion in online social networks. *Eur. Phys. J. B* 86, 1 (2013), 1–7.

Alton Y. K. Chua and Snehasish Banerjee. 2016. Linguistic predictors of rumor veracity on the internet. In *Proceedings of the International MultiConference of Engineers and Computer Scientists*, Vol. 1.

Alton Y. K. Chua, Cheng-Ying Tee, Augustine Pang, and Ee-Peng Lim. 2016. The retransmission of rumor-related tweets: Characteristics of source and message. In *Proceedings of the 7th 2016 International Conference on Social Media & Society*. ACM, 22.

Anh Dang, Abidalrahman Moh'd, Evangelos Milios, Rosane Minghim, others et al. 2016a. What is in a rumour: Combined visual analysis of rumour flow and user activity. *Comput. Graph. Int., XXXIII* (2016).

Anh Dang, Michael Smit, Abidalrahman Moh'd, Rosane Minghim, and Evangelos Milios. 2016b. Toward understanding how users respond to rumours in social media. In *Proceedings of the 2016 IEEE/ACM International Conference on Advances in Social Networks Analysis and Mining (ASONAM'16)*. IEEE, 777–784.

Thierry Declerck, Petya Osenova, Georgi Georgiev, and Piroska Lendvai. 2015. Ontological modelling of rumors. In *Proceedings of the Workshop on Social Media and the Web of Linked Data*. Springer, 3–17.

Leon Derczynski and Kalina Bontcheva. 2014. Pheme: Veracity in digital social networks. In *UMAP Workshops*.

Leon Derczynski, Kalina Bontcheva, Maria Liakata, Rob Procter, Geraldine Wong Sak Hoi, and Arkaitz Zubiaga. 2017. SemEval-2017 Task 8: RumourEval: Determining rumour veracity and support for rumours. In *Proceedings of SemEval*. ACL.

Nicholas Diakopoulos, Munmun De Choudhury, and Mor Naaman. 2012. Finding and assessing social media information sources in the context of journalism. In *Proceedings of the SIGCHI Conference on Human Factors in Computing Systems*. ACM, 2451–2460.

Nicholas DiFonzo and Prashant Bordia. 2007. Rumor, gossip and urban legends. *Diogenes* 54, 1 (2007), 19–35.

Anhai Doan, Raghu Ramakrishnan, and Alon Y. Halevy. 2011. Crowdsourcing systems on the world-wide web. *Commun. ACM* 54, 4 (2011), 86–96.

Pamela Donovan. 2007. How idle is idle talk? One hundred years of rumor research. *Diogenes* 54, 1 (2007), 59–82.

Kevin Driscoll and Shawn Walker. 2014. Big data, big questions| working within a black box: Transparency in the collection and production of big twitter data. *International Journal of Communication* 8 (2014), 20.

Paul S. Earle, Daniel C. Bowden, and Michelle Guy. 2012. Twitter earthquake detection: Earthquake monitoring in a social world. *Ann. Geophys.* 54, 6 (2012).

Omar Enayet and Samhaa R. El-Beltagy. 2017. NileTMRG at SemEval-2017 Task 8: Determining rumour and veracity support for rumours on Twitter. In *Proceedings of SemEval*. ACL.

William Ferreira and Andreas Vlachos. 2016. Emergent: A novel data-set for stance classification. In *Proceedings of the 2016 Conference of the North American Chapter of the Association for Computational Linguistics: Human Language Technologies*. ACL.

Samantha Finn, Panagiotis Takis Metaxas, and Eni Mustafaraj. 2015. Spread and skepticism: Metrics of propagation on twitter. In *Proceedings of the ACM Web Science Conference*. ACM, 39.

Samantha Finn, Panagiotis Takis Metaxas, Eni Mustafaraj, Megan O'Keefe, Lindsay Tang, Susan Tang, and Laura Zeng. 2014. TRAILS: A system for monitoring the propagation of rumors on Twitter. In *Computation and Journalism Symposium, NYC, NY*.

Víctor Fresno, Arkaitz Zubiaga, Heng Ji, and Raquel Martínez. 2015. Exploiting geolocation, user and temporal information for natural hazards monitoring in Twitter. *Proces. Leng. Nat.* 54 (2015), 85–92.

Vanessa Frias-Martinez, Victor Soto, Heath Hohwald, and Enrique Frias-Martinez. 2012. Characterizing urban landscapes using geolocated tweets. In *International Conference on Privacy, Security, Risk and Trust and International Conference on Social Computing*. IEEE, 239–248.

Christian Fuchs. 2013. *Social Media: A Critical Introduction.* Sage.

Yue Gao, Fanglin Wang, Huanbo Luan, and Tat-Seng Chua. 2014. Brand data gathering from live social media streams. In *Proceedings of International Conference on Multimedia Retrieval*. ACM, 169.

Marianela García Lozano, Hanna Lilja, Edward Tjörnhammar, and Maja Maja Karasalo. 2017. Mama Edha at SemEval-2017 Task 8: Stance classification with CNN and rules. In *Proceedings of SemEval*. ACL.

Georgios Giasemidis, Colin Singleton, Ioannis Agrafiotis, Jason R. C. Nurse, Alan Pilgrim, Chris Willis, and D.V. Greetham. 2016. Determining the veracity of rumours on Twitter. In *International Conference on Social Informatics*. Springer, 185–205.

Michael B. Goodman, Norman Booth, and Julie Ann Matic. 2011. Mapping and leveraging influencers in social media to shape corporate brand perceptions. *Corp. Commun.: Int. J.* 16, 3 (2011), 184–191.

Jeffrey Gottfried and Elisa Shearer. 2016. *News Use Across Social Media Platforms 2016.* Technical Report. Pew Research Center.

Weiwei Guo and Mona Diab. 2012. Modeling sentences in the latent space. In *Proceedings of the 50th Annual Meeting of the Association for Computational Linguistics: Long Papers.*, Vol. 1. Association for Computational Linguistics, 864–872.

Aditi Gupta and Ponnurangam Kumaraguru. 2012. Credibility ranking of tweets during high impact events. In *Proceedings of the 1st Workshop on Privacy and Security in Online Social Media*. ACM, 2.

Aditi Gupta, Ponnurangam Kumaraguru, Carlos Castillo, and Patrick Meier. 2014. Tweetcred: Real-time credibility assessment of content on twitter. In *Proceedings of the International Conference on Social Informatics*. Springer, 228–243.

Mark Hall, Eibe Frank, Geoffrey Holmes, Bernhard Pfahringer, Peter Reutemann, and Ian H. Witten. 2009. The WEKA data mining software: An update. *ACM SIGKDD Explor. Newsl.* 11, 1 (2009), 10–18.

Sardar Hamidian and Mona T. Diab. 2015. Rumor detection and classification for twitter data. In *Proceedings of the 5th International Conference on Social Media Technologies, Communication, and Informatics (SOTICS'15)*. IARIA, 71–77.

Sardar Hamidian and Mona T. Diab. 2016. Rumor identification and belief investigation on twitter. In *Proceedings of NAACL-HLT*. 3–8. Association for Computational Linguistics.

Bo Han, Paul Cook, and Timothy Baldwin. 2014. Text-based twitter user geolocation prediction. *J. Artif. Intell. Res.* 49 (2014), 451–500.

Aniko Hannak, Drew Margolin, Brian Keegan, and Ingmar Weber. 2014. Get back! you dont know me like that: The social mediation of fact checking interventions in twitter conversations. In *Proceedings of the AAAI Conference on Weblogs and Social Media*.

Naeemul Hassan, Bill Adair, James T. Hamilton, Chengkai Li, Mark Tremayne, Jun Yang, and Cong Yu. 2015. The quest to automate fact-checking. In *Proceedings of the Computation and Journalism Symposium, NYC, NY*.

Alfred Hermida. 2010. Twittering the news: The emergence of ambient journalism. *Journal. Pract.* 4, 3 (2010), 297–308.

Alfred Hermida. 2012. Tweets and truth: Journalism as a discipline of collaborative verification. *Journal. Pract.* 6, 5–6 (2012), 659–668.

Alfred Hermida and Neil Thurman. 2008. A clash of cultures: The integration of user-generated content within professional journalistic frameworks at British newspaper websites. *Journal. Pract.* 2, 3 (2008), 343–356.

Peter Hernon. 1995. Disinformation and misinformation through the Internet: Findings of an exploratory study. *Gov. Inform. Q.* 12, 2 (1995), 133–139.

Elle Hunt. 2016. What is fake news? How to spot it and what you can do to stop it. *The Guardian* (December 2016). https://www.theguardian.com/media/2016/dec/18/what-is-fake-news-pizzagate.

Muhammad Imran, Carlos Castillo, Fernando Diaz, and Sarah Vieweg. 2015. Processing social media messages in mass emergency: A survey. *ACM Computing Surveys (CSUR)* 47, 4 (2015), 67.

Marianne E. Jaeger, Susan Anthony, and Ralph L. Rosnow. 1980. Who hears what from whom and with what effect a study of rumor. *Personal. Soc. Psychol. Bull.* 6, 3 (1980), 473–478.

Kokil Jaidka, Kaushik Ramachandran, Prakhar Gupta, and Sajal Rustagi. 2016. SocialStories: Segmenting stories within trending twitter topics. In *Proceedings of the 3rd IKDD Conference on Data Science, 2016*. ACM, 1.

Dong-Gi Jo. 2002. Diffusion of rumors on the internet. *Inform. Soc. Rev.* 2002 (2002), 77–95.

Robert H. Knapp. 1944. A psychology of rumor. *Public Opin. Q.* 8, 1 (1944), 22–37.

Elena Kochkina, Maria Liakata, and Isabelle Augenstein. 2017. Turing at SemEval-2017 Task 8: Sequential approach to rumour stance classification with branch-LSTM. In *Proceedings of SemEval*. ACL.

Haewoon Kwak, Changhyun Lee, Hosung Park, and Sue Moon. 2010. What is Twitter, a social network or a news media? In *Proceedings of the 19th International Conference on World Wide Web*. ACM, 591–600.

Sejeong Kwon and Meeyoung Cha. 2014. Modeling bursty temporal pattern of rumors. In *Proceedings of the International Conference on Weblogs and Social Media (ICWSM)*. AAAI, 650–651.

Sejeong Kwon, Meeyoung Cha, and Kyomin Jung. 2017. Rumor detection over varying time windows. *PLOS ONE* 12, 1 (2017), e0168344.

Sejeong Kwon, Meeyoung Cha, Kyomin Jung, Wei Chen, and Yajun Wang. 2013. Prominent features of rumor propagation in online social media. In *2013 IEEE 13th International Conference on Data Mining*. IEEE, 1103–1108.

David Lazer, Alex Sandy Pentland, Lada Adamic, Sinan Aral, Albert Laszlo Barabasi, Devon Brewer, Nicholas Christakis, Noshir Contractor, James Fowler, Myron Gutmann, et al. 2009. Life in the network: The coming age of computational social science. *Science* 323, 5915 (2009), 721.

Lian Fen Lee, Amy P. Hutton, and Susan Shu. 2015. The role of social media in the capital market: Evidence from consumer product recalls. *J. Account. Res.* 53, 2 (2015), 367–404.

Piroska Lendvai, Isabelle Augenstein, Kalina Bontcheva, and Thierry Declerck. 2016a. Monolingual social media datasets for detecting contradiction and entailment. In *Proceedings of the 10th International Conference on Language Resources and Evaluation (LREC), Portoroz, Slovenia. European Language Resources Association (ELRA'16)*.

Piroska Lendvai and Uwe D. Reichel. 2016. Contradiction detection for rumorous claims. In *Proceedings of the Workshop on Extra-Propositional Aspects of Meaning in Computational Linguistics*. 31–40.

Piroska Lendvai, Uwe D. Reichel, and Thierry Declerck. 2016b. Factuality drift assessment by lexical markers in resolved rumors. In *Joint Proceedings of the Posters and Demos Track of the 12th International Conference on Semantic Systems (SEMANTiCS'16) and the 1st International Workshop on Semantic Change & Evolving Semantics, Leipzig*. ACM. http://ceurws.org/Vol-1695/.

Stephan Lewandowsky, Ullrich K. H. Ecker, Colleen M. Seifert, Norbert Schwarz, and John Cook. 2012. Misinformation and its correction continued influence and successful debiasing. *Psychol. Sci. Public Interest* 13, 3 (2012), 106–131.

Huaye Li and Yasuaki Sakamoto. 2015. Computing the veracity of information through crowds: A method for reducing the spread of false messages on social media. In *Proceedings of the 2015 48th Hawaii International Conference onSystem Sciences (HICSS'15)*. IEEE, 2003–2012.

Jiwei Li and Claire Cardie. 2014. Timeline generation: Tracking individuals on Twitter. In *Proceedings of the 23rd International Conference on World Wide Web*. ACM, 643–652.

Gang Liang, Wenbo He, Chun Xu, Liangyin Chen, and Jinquan Zeng. 2015. Rumor identification in microblogging systems based on users behavior. *IEEE Trans. Computat. Soc. Syst.* 2, 3 (2015), 99–108.

Qinying Liao and Lei Shi. 2013. She gets a sports car from our donation: Rumor transmission in a chinese microblogging community. In *Proceedings of the 2013 Conference on Computer Supported Cooperative Work*. ACM, 587–598.

Fang Liu, Andrew Burton-Jones, and Dongming Xu. 2014. Rumors on social media in disasters: Extending transmission to retransmission. In *PACIS*. 49.

Xiaomo Liu, Armineh Nourbakhsh, Quanzhi Li, Rui Fang, and Sameena Shah. 2015. Real-time rumor debunking on Twitter. In *Proceedings of the 24th ACM International on Conference on Information and Knowledge Management*. ACM, 1867–1870.

Stine Lomborg and Anja Bechmann. 2014. Using APIs for data collection on social media. *Inform. Soc.* 30, 4 (2014), 256–265.

Rui Long, Haofen Wang, Yuqiang Chen, Ou Jin, and Yong Yu. 2011. Towards effective event detection, tracking and summarization on microblog data. In *International Conference on Web-Age Information Management*. Springer, 652–663.

Michal Lukasik, Trevor Cohn, and Kalina Bontcheva. 2015a. Classifying tweet level judgements of rumours in social media. In *Proceedings of the 2015 Conference on Empirical Methods in Natural Language Processing (EMNLP'15)*. 2590–2595.

Michal Lukasik, Trevor Cohn, and Kalina Bontcheva. 2015b. Point process modelling of rumour dynamics in social media. In *ACL (2)*. 518–523.

Michal Lukasik, P. K. Srijith, Duy Vu, Kalina Bontcheva, Arkaitz Zubiaga, and Trevor Cohn. 2016. Hawkes processes for continuous time sequence classification: An application to rumour stance classification in Twitter. In *Proceedings of 54th Annual Meeting of the Association for Computational Linguistics*. Association for Computational Linguistics, 393–398.

Ben Ma, Dazhen Lin, and Donglin Cao. 2017. Content representation for microblog rumor detection. In *Advances in Computational Intelligence Systems*. Springer, 245–251.

Jing Ma, Wei Gao, Zhongyu Wei, Yueming Lu, and Kam-Fai Wong. 2015. Detect rumors using time series of social context information on microblogging websites. In *Proceedings of the 24th ACM International on Conference on Information and Knowledge Management*. ACM, 1751–1754.

Jim Maddock, Kate Starbird, Haneen J. Al-Hassani, Daniel E. Sandoval, Mania Orand, and Robert M. Mason. 2015. Characterizing online rumoring behavior using multi-dimensional signatures. In *Proceedings of the 18th ACM Conference on Computer Supported Cooperative Work & Social Computing*. ACM, 228–241.

Adam Marcus, Michael S. Bernstein, Osama Badar, David R. Karger, Samuel Madden, and Robert C. Miller. 2011. Twitinfo: Aggregating and visualizing microblogs for event exploration. In *Proceedings of the SIGCHI Conference on Human Factors in Computing Systems*. ACM, 227–236.

Richard McCreadie, Craig Macdonald, and Iadh Ounis. 2015. Crowdsourced rumour identification during emergencies. In *Proceedings of the 24th International Conference on World Wide Web*. ACM, 965–970.

Marcelo Mendoza, Barbara Poblete, and Carlos Castillo. 2010. Twitter under crisis: Can we trust what we RT? In *Proceedings of the First Workshop on Social Media Analytics*. ACM, 71–79.

Panagiotis Takas Metaxas, Samantha Finn, and Eni Mustafaraj. 2015. Using Twittertrails.com to investigate rumor propagation. In *Proceedings of the 18th ACM Conference Companion on Computer Supported Cooperative Work & Social Computing*. ACM, 69–72.

Stuart E. Middleton and Vadims Krivcovs. 2016. Geoparsing and geosemantics for social media: Spatio-temporal grounding of content propagating rumours to support trust and veracity analysis during breaking news. *ACM Trans. Inform. Syst.* 34, 3 (2016), 1–27.

Stuart E. Middleton, Lee Middleton, and Stefano Modafferi. 2014. Real-time crisis mapping of natural disasters using social media. *IEEE Intell. Syst.* 29, 2 (2014), 9–17.

George A. Miller. 1995. WordNet: A lexical database for english. *Commun. ACM* 38, 11 (1995), 39–41.

Tanushree Mitra and Eric Gilbert. 2015. CREDBANK: A large-scale social media corpus with associated credibility annotations. In *ICWSM*. 258–267.

Saif M. Mohammad, Svetlana Kiritchenko, Parinaz Sobhani, Xiaodan Zhu, and Colin Cherry. 2016. Semeval-2016 Task 6: Detecting stance in tweets. In *Proceedings of the International Workshop on Semantic Evaluation, SemEval*. Association for Computational Linguistics, 31–41.

Joe Murphy, Michael W. Link, Jennifer Hunter Childs, Casey Langer Tesfaye, Elizabeth Dean, Michael Stern, Josh Pasek, Jon Cohen, Mario Callegaro, and Paul Harwood. 2014. Social media in public opinion research executive summary of the aapor task force on emerging technologies in public opinion research. *Public Opin. Q.* 78, 4 (2014), 788–794.

Robert Nares. 1822. *A Glossary: Or, Collection of Words, Phrases, Names, and Allusions to Customs, Proverbs, Etc., which Have Been Thought to Require Illustration, in the Works of English Authors, Particularly Shakespeare, and His Contemporaries*. R. Triphook.

Onook Oh, Manish Agrawal, and H. Raghav Rao. 2013. Community intelligence and social media services: A rumor theoretic analysis of tweets during social crises. *Mis Q.* 37, 2 (2013), 407–426.

Alexandra Olteanu, Carlos Castillo, Fernando Diaz, and Emre Kiciman. 2016. Social data: Biases, methodological pitfalls, and ethical boundaries. (2016).

Symeon Papadopoulos, Kalina Bontcheva, Eva Jaho, Mihai Lupu, and Carlos Castillo. 2016. Overview of the special issue on trust and veracity of information in social media. *ACM Trans. Inform. Syst. (TOIS'16)* 34, 3 (2016), 14.

Richard Petty and John Cacioppo. 2012. *Communication and Persuasion: Central and Peripheral Routes to Attitude Change*. Springer Science & Business Media.

Swit Phuvipadawat and Tsuyoshi Murata. 2010. Breaking news detection and tracking in Twitter. In *Proceedings of the 2010 IEEE/WIC/ACM International Conference on Web Intelligence and Intelligent Agent Technology (WI-IAT'10)*, Vol. 3. IEEE, 120–123.

Kashyap Popat, Subhabrata Mukherjee, Jannik Strötgen, and Gerhard Weikum. 2016. Credibility assessment of textual claims on the web. In *Proceedings of the 25th ACM International on Conference on Information and Knowledge Management*. ACM, 2173–2178.

Jamuna Prasad. 1935. The psychology of rumour: A study relating to the great indian earthquake of 1934. *Br. J. Psychol. Gen. Sect.* 26, 1 (1935), 1–15.

Rob Procter, Jeremy Crump, Susanne Karstedt, Alex Voss, and Marta Cantijoch. 2013a. Reading the riots: What were the police doing on Twitter? *Polic. Soc.* 23, 4 (2013), 413–436.

Rob Procter, Farida Vis, and Alex Voss. 2013b. Reading the riots on Twitter: Methodological innovation for the analysis of big data. *Int. J. Soc. Res. Methodol.* 16, 3 (2013), 197–214.

Vahed Qazvinian, Emily Rosengren, Dragomir R. Radev, and Qiaozhu Mei. 2011. Rumor has it: Identifying misinformation in microblogs. In *Proceedings of EMNLP*. 1589–1599.

Yumeng Qin, Dominik Wurzer, Victor Lavrenko, and Cunchen Tang. 2016. Spotting rumors via novelty detection. *arXiv:1611.06322* (2016).

Paul Resnick, Samuel Carton, Souneil Park, Yuncheng Shen, and Nicole Zeffer. 2014. Rumorlens: A system for analyzing the impact of rumors and corrections in social media. In *Proceedings of the Computational Journalism Conference*.

Timo Reuter and Philipp Cimiano. 2012. Event-based classification of social media streams. In *Proceedings of the 2nd ACM International Conference on Multimedia Retrieval*. ACM, 22.

Maarten de Rijke. 2016. Special issue on trust and veracity of information in social media. *ACM Trans. Inform. Syst.* 34, 3 (2016).

Sara Rosenthal and Kathleen McKeown. 2015. I couldn't agree more: The role of conversational structure in agreement and disagreement detection in online discussions. In *Proceedings of the 16th Annual Meeting of the Special Interest Group on Discourse and Dialogue*. 168.

Ralph L. Rosnow and Eric K. Foster. 2005. Rumor and gossip research. *Psychol. Sci. Agenda* 19, 4 (2005).

Hassan Sayyadi, Matthew Hurst, and Alexey Maykov. 2009. Event detection and tracking in social streams. In *Proceedings of the AAAI Conference on Weblogs and Social Media*.

T. Joseph Scanlon. 1977. Post-disaster rumor chains: A case study. *Mass Emerg.* 2, 126 (1977), 22–27.

Salvatore Scifo and Lemi Baruh. 2015. *D2.3. Report on the Adverse Use and Reliability of New Media*. Technical Report. COSMIC Project Deliverable.

Eunsoo Seo, Prasant Mohapatra, and Tarek Abdelzaher. 2012. Identifying rumors and their sources in social networks. In *Proceedings of the Conference on SPIE Defense, Security, and Sensing*. International Society for Optics and Photonics, 83891I–83891I.

Emilio Serrano, Carlos A. Iglesias, and Mercedes Garijo. 2015. A survey of Twitter rumor spreading simulations. In *Comput. Collect. Intell.* Springer, 113–122.

Chengcheng Shao, Giovanni Luca Ciampaglia, Alessandro Flammini, and Filippo Menczer. 2016. Hoaxy: A platform for tracking online misinformation. In *Proceedings of the 25th International Conference Companion on World Wide Web*. 745–750.

Rui Shi, Paul Messaris, and Joseph N. Cappella. 2014. Effects of online comments on smokers' perception of antismoking public service announcements. *J. Comput.-Med. Commun.* 19, 4 (2014), 975–990.

Jieun Shin, Lian Jian, Kevin Driscoll, and François Bar. 2016. Political rumoring on Twitter during the 2012 US presidential election: Rumor diffusion and correction. *New Media & Society* (2016), 1461444816634054.

Craig Silverman. 2013. *Verification Handbook*. The European Journalism Centre.

Craig Silverman. 2015a. *Lies, Damn Lies, and Viral Content: How News Websites Spread (and Debunk) Online Rumors, Unverified Claims and Misinformation*. Tow Center for Digital Journalism.

Craig Silverman. 2015b. *Verification Handbook for Investigative Reporting: A Guide to Online Search and Research Techniques for Using UGC and Open Source Information in Investigations*. European Journalism Centre.

Vikram Singh, Sunny Narayan, Md Shad Akhtar, Asif Ekbal, and Pushpak Bhattacharya. 2017. IITP at SemEval-2017 Task 8: A supervised approach for rumour evaluation. In *Proceedings of SemEval*. Association for Computational Linguistics, 497–501.

Durganand Sinha. 1952. Behaviour in a catastrophic situation: A psychological study of reports and rumours. *Br. J. Psycho. Gen. Sect.* 43, 3 (1952), 200–209.

Swapna Somasundaran and Janyce Wiebe. 2009. Recognizing stances in online debates. In *Proceedings of the Joint Conference of the 47th Annual Meeting of the ACL and the 4th International Joint Conference on Natural Language Processing of the AFNLP, Vol. 1.*. Association for Computational Linguistics, 226–234.

Jochen Spangenberg and Nicolaus Heise. 2014. News from the crowd: Grassroots and collaborative journalism in the digital age. In *Proceedings of the 23rd International Conference on World Wide Web*. 765–768.

Ankit Srivastava, Rehm Rehm, and Julian Moreno Schneider. 2017. DFKI-DKT at SemEval-2017 Task 8: Rumour detection and classification using cascading heuristics. In *Proceedings of SemEval*. ACL, 486–490.

Kate Starbird, Jim Maddock, Mania Orand, Peg Achterman, and Robert M. Mason. 2014. Rumors, false flags, and digital vigilantes: Misinformation on Twitter after the 2013 boston marathon bombing. *iConference 2014 Proceedings* (2014). iSchools, 654–662.

Kate Starbird, Grace Muzny, and Leysia Palen. 2012. Learning from the crowd: Collaborative filtering techniques for identifying on-the-ground Twitterers during mass disruptions. In *Proceedings of 9th International Conference on Information Systems for Crisis Response and Management, (ISCRAM'12)*.

Kang Hoon Sung and Moon J. Lee. 2015. Do online comments influence the public's attitudes toward an organization? Effects of online comments based on individuals prior attitudes. *J. Psychol* 149, 4 (2015), 325–338.

Tetsuro Takahashi and Nobuyuki Igata. 2012. Rumor detection on Twitter. In *Proceedings of the 2012 Joint 6th International Conference on Soft Computing and Intelligent Systems (SCIS'12) and 13th International Symposium on Advanced Intelligent Systems (ISIS'12)*. IEEE, 452–457.

Misako Takayasu, Kazuya Sato, Yukie Sano, Kenta Yamada, Wataru Miura, and Hideki Takayasu. 2015. Rumor diffusion and convergence during the 3.11 earthquake: A Twitter case study. *PLoS One* 10, 4 (2015), e0121443.

Yla R. Tausczik and James W. Pennebaker. 2010. The psychological meaning of words: LIWC and computerized text analysis methods. *J. Lang. Soc. Psychol.* 29, 1 (2010), 24–54.

Peter Tolmie, Rob Procter, Dave Randall, Mark Rouncefield, Christian Burger, Geraldine Wong Sak Hoi, Arkaitz Zubiaga, and Maria Liakata. 2017. Supporting the use of user generated content in journalistic practice. In *Proceedings of the ACM Conference on Human Factors and Computing Systems*.

Peter Tolmie, Rob Procter, Mark Rouncefield, Maria Liakata, and Arkaitz Zubiaga. 2018. Microblog analysis as a programme of work. *ACM Transactions on Social Computing* 1, 1 (2018), Article 2, 40. DOI: https://doi.org/10.1145/3162956

Laura Tolosi, Andrey Tagarev, and Georgi Georgiev. 2016. An analysis of event-agnostic features for rumour classification in Twitter. In *Proceedings of the ICWSM Workshop on Social Media in the Newsroom*.

Guangmo Tong, Weili Wu, Ling Guo, Deying Li, Cong Liu, Bin Liu, and Ding-Zhu Du. 2017. An efficient randomized algorithm for rumor blocking in online social networks. arXiv:1701.02368 (2017).

Daniel Trottier and Christian Fuchs. 2014. *Social Media, Politics and the State: Protests, Revolutions, Riots, Crime and Policing in the Age of Facebook, Twitter and YouTube*, Vol. 16. Routledge.

Christos Tzelepis, Zhigang Ma, Vasileios Mezaris, Bogdan Ionescu, Ioannis Kompatsiaris, Giulia Boato, Nicu Sebe, and Shuicheng Yan. 2016. Event-based media processing and analysis: A survey of the literature. *Image Vis. Comput.* 53 (2016), 3–19.

José Van Dijck. 2013. *The Culture of Connectivity: A Critical History of Social Media*. Oxford University Press.

Sarah Vieweg, Amanda L. Hughes, Kate Starbird, and Leysia Palen. 2010. Microblogging during two natural hazards events: What Twitter may contribute to situational awareness. In *Proceedings of the SIGCHI Conference on Human Factors in Computing Systems*. ACM, 1079–1088.

Soroush Vosoughi. 2015. *Automatic Detection and Verification of Rumors on Twitter*. Ph.D. Dissertation.

Rhythm Walia and M. P. S. Bhatia. 2016. Modeling rumors in Twitter: An overview. *Int. J. Rough Sets Data Anal. (IJRSDA)* 3, 4 (2016), 46–67.

Marilyn A. Walker, Pranav Anand, Rob Abbott, Jean E. Fox Tree, Craig Martell, and Joseph King. 2012. That is your evidence? Classifying stance in online political debate. *Decis. Support Syst.* 53, 4 (2012), 719–729.

Aobo Wang, Cong Duy Vu Hoang, and Min-Yen Kan. 2013. Perspectives on crowdsourcing annotations for natural language processing. *Lang. Resou. Eval.* 47, 1 (2013), 9–31.

Feixiang Wang, Man Lan, and Yuanbin Wu. 2017. ECNU at SemEval-2017 Task 8: Rumour evaluation using effective features and supervised ensemble models. In *Proceedings of SemEval*. ACL, 491–496.

Shihan Wang and Takao Terano. 2015. Detecting rumor patterns in streaming social media. In *Proceedings of the 2015 IEEE International Conference on Big Data (Big Data'15)*. IEEE, 2709–2715.

Wen Wang, Sibel Yaman, Kristin Precoda, Colleen Richey, and Geoffrey Raymond. 2011. Detection of agreement and disagreement in broadcast conversations. In *Proceedings of the 49th Annual Meeting of the Association for Computational Linguistics: Human Language Technologies: Short Papers, Vol. 2*. Association for Computational Linguistics, 374–378.

Xinyue Wang, Laurissa Tokarchuk, Felix Cuadrado, and Stefan Poslad. 2015. Adaptive identification of hashtags for real-time event data collection. In *Recommendation and Search in Social Networks*. Springer, 1–22.

Claire Wardle. 2017. Fake news. It's complicated. *First Draft News* (February 2017). https://firstdraftnews.org/fake-news-complicated/.

H. Webb, P. Burnap, R. Procter, O. Rana, B. C. Stahl, M. Williams, W. Housley, A. Edwards, and M. Jirotka. 2016. Digital wildfires: Propagation, verification, regulation, and responsible innovation. *ACM Trans. Inform. Syst.* 34, 3 (2016).

Katrin Weller and Katharina E. Kinder-Kurlanda. 2016. A manifesto for data sharing in social media research. In *Proceedings of the 8th ACM Conference on Web Science*. ACM, 166–172.

David Westerman, Patric R. Spence, and Brandon Van Der Heide. 2014. Social media as information source: Recency of updates and credibility of information. *J. Comput.-Med. Commun.* 19, 2 (2014), 171–183.

Ke Wu, Song Yang, and Kenny Q. Zhu. 2015. False rumors detection on sina weibo by propagation structures. In *Proceedings of the 2015 IEEE 31st International Conference on Data Engineering*. IEEE, 651–662.

Fan Yang, Yang Liu, Xiaohui Yu, and Min Yang. 2012. Automatic detection of rumor on Sina Weibo. In *Proceedings of the ACM SIGKDD Workshop on Mining Data Semantics*. ACM, 13.

YeKang Yang, Kai Niu, and ZhiQiang He. 2015. Exploiting the topology property of social network for rumor detection. In *Proceedings of the 2015 12th International Joint Conference on Computer Science and Software Engineering (JCSSE'15)*. IEEE, 41–46.

Dave Yates and Scott Paquette. 2011. Emergency knowledge management and social media technologies: A case study of the 2010 Haitian earthquake. *Int. J. Inform. Manag.* 31, 1 (2011), 6–13.

Jie Yin, Andrew Lampert, Mark Cameron, Bella Robinson, and Robert Power. 2012. Using social media to enhance emergency situation awareness. *IEEE Intell. Syst.* 27, 6 (2012), 52–59.

Li Zeng, Kate Starbird, and Emma S. Spiro. 2016. #Unconfirmed: Classifying rumor stance in crisis-related social media messages. In *Proceedings of the Tenth International AAAI Conference on Web and Social Media*. AAAI, 747–750.

Zili Zhang, Ziqiong Zhang, and Hengyun Li. 2015. Predictors of the authenticity of internet health rumours. *Health Inform. Libr. J.* 32, 3 (2015), 195–205.

Zhe Zhao, Paul Resnick, and Qiaozhu Mei. 2015. Enquiring minds: Early detection of rumors in social media from enquiry posts. In *Proceedings of the 24th International Conference on World Wide Web*. ACM, 1395–1405.

Arkaitz Zubiaga and Heng Ji. 2014. Tweet, but verify: Epistemic study of information verification on Twitter. *Soc. Netw. Anal. Min.* 4, 1 (2014), 1–12.

Arkaitz Zubiaga, Heng Ji, and Kevin Knight. 2013. Curating and contextualizing Twitter stories to assist with social news-gathering. In *Proceedings of the 2013 International Conference on Intelligent User Interfaces*. ACM, 213–224.

Arkaitz Zubiaga, Elena Kochkina, Maria Liakata, Rob Procter, and Michal Lukasik. 2016a. Stance classification in rumours as a sequential task exploiting the tree structure of social media conversations. In *Proceedings of the 26th International Conference on Computational Linguistics*. Association for Computational Linguistics, 2438–2448.

Arkaitz Zubiaga, Maria Liakata, and Rob Procter. 2016b. Learning reporting dynamics during breaking news for rumour detection in social media. *arXiv:1610.07363* (2016).

Arkaitz Zubiaga, Maria Liakata, and Rob Procter. 2017. Exploiting context for rumour detection in social media. In *Proceedings of the International Conference on Social Informatics*. Springer, 109–123.

Arkaitz Zubiaga, Maria Liakata, Rob Procter, Kalina Bontcheva, and Peter Tolmie. 2015. Crowdsourcing the annotation of rumourous conversations in social media. In *Proceedings of the 24th International Conference on World Wide Web Companion*. ACM, 347–353.

Arkaitz Zubiaga, Maria Liakata, Rob Procter, Geraldine Wong Sak Hoi, and Peter Tolmie. 2016c. Analysing how people orient to and spread rumours in social media by looking at conversational threads. *PLoS ONE* 11, 3 (03 2016), 1–29. DOI:http://dx.doi.org/10.1371/journal.pone.0150989